\RequirePackage{fix-cm}
\documentclass[twocolumn,compsoc]{CVM}

\graphicspath{{figures/},{figures/cvm/}}

\usepackage{amsmath}
\usepackage{graphicx}
\usepackage{multirow}
\usepackage{subfigure}
\usepackage{caption}
\usepackage{rotating}
\usepackage{cite}
\usepackage{amssymb}
\usepackage{array}
\usepackage{bm}
\usepackage{color}
\usepackage{enumitem}
\usepackage{overpic}
\usepackage{colortbl}
\usepackage{subcaption}
\usepackage{multicol}

\usepackage[export]{adjustbox} %
\usepackage{tabularx} %
\usepackage{booktabs}  %
\usepackage[ruled,longend,noline]{algorithm2e}
\usepackage{makecell}

\let\Letter\relax
\usepackage{marvosym}
\usepackage{pifont}

\definecolor{green}{RGB}{0,255,0} 
\definecolor{red}{RGB}{255,0,0} 
\usepackage[colorlinks,
            linkcolor=red,
            anchorcolor=blue,
            citecolor=green]{hyperref}

\usepackage{hyperref}
\makeatletter
\def\UrlAlphabet{%
      \do\a\do\b\do\c\do\d\do\e\do\f\do\g\do\h\do\i\do\j%
      \do\k\do\l\do\m\do\n\do\o\do\p\do\q\do\r\do\s\do\t%
      \do\u\do\v\do\w\do\x\do\y\do\z\do\A\do\B\do\C\do\D%
      \do\E\do\F\do\G\do\H\do\I\do\J\do\K\do\L\do\M\do\N%
      \do\O\do\P\do\Q\do\R\do\S\do\T\do\U\do\V\do\W\do\X%
      \do\Y\do\Z}
\def\UrlDigits{\do\1\do\2\do\3\do\4\do\5\do\6\do\7\do\8\do\9\do\0}
\g@addto@macro{\UrlBreaks}{\UrlOrds}
\g@addto@macro{\UrlBreaks}{\UrlAlphabet}
\g@addto@macro{\UrlBreaks}{\UrlDigits}
\makeatother

\definecolor{mygray}{gray}{.90}

\def\ManuscriptInfo{Manuscript received: 2024-06-12; accepted: 2024-09-30}

\def\PaperTitle{StyleDiffusion: Prompt-Embedding Inversion for Text-Based Editing}

\def\AuthorNames{
Senmao Li$^{1}$, Joost van de Weijer$^{2}$, Taihang Hu$^{1}$, Fahad Shahbaz Khan$^{3}$, Qibin Hou$^{1}$, \\Yaxing Wang$^{1}$(\Letter), Jian Yang$^{1}$, Ming-Ming Cheng$^{1}$}

\def\AuthorAdress{
$1\quad $ & College of Computer Science, Nankai University. Tianjin, China 300350.  \\
$2\quad $ & The Computer Vision Center, Universitat Aut\`onoma de Barcelona, Barcelona 08193, Spain. E-mail: joost@cvc.uab.es.\\
$3\quad $ & Mohamed bin Zayed University of Artificial Intelligence (UAE),  and Linkoping University (Sweden). E-mail: fahad.khan@liu.se. \\
}

\def\firstheader{https://doi.org/10.26599/CVM.2025.9450462\\[5mm]~Short Communication}

\headevenname{S. Li, J. van de Weijer, T. Hu, et al.}

\begin{document}

\MakePageStyle

\MakeAbstract{
A significant research effort is focused on exploiting the amazing capacities of pretrained diffusion models for the editing of images.They either finetune the model, or invert the image in the latent space of the pretrained model. However, they suffer from two problems: (1) Unsatisfying results for selected regions and unexpected changes in non-selected regions.
(2) They require careful text prompt editing where the prompt should include all visual objects in the input image.To address this, we propose two improvements: (1) Only optimizing the input of the value linear network in the cross-attention layers is sufficiently powerful to reconstruct a real image. (2) We propose attention regularization to preserve the object-like attention maps after reconstruction and editing, enabling us to obtain accurate style editing without invoking significant structural changes. We further improve the editing technique that is used for the unconditional branch of classifier-free guidance as used by P2P. Extensive experimental prompt-editing results on a variety of images demonstrate qualitatively and quantitatively that our method has superior editing capabilities compared to existing and concurrent works. See our accompanying code in  Stylediffusion: \url{https://github.com/sen-mao/StyleDiffusion}.
}

\MakeKeywords{Real-image inversion, Image editing, Diffusion models}

\section{Introduction}
\label{intro}

\begin{figure*}[t]
\centering
\includegraphics[width=\linewidth]{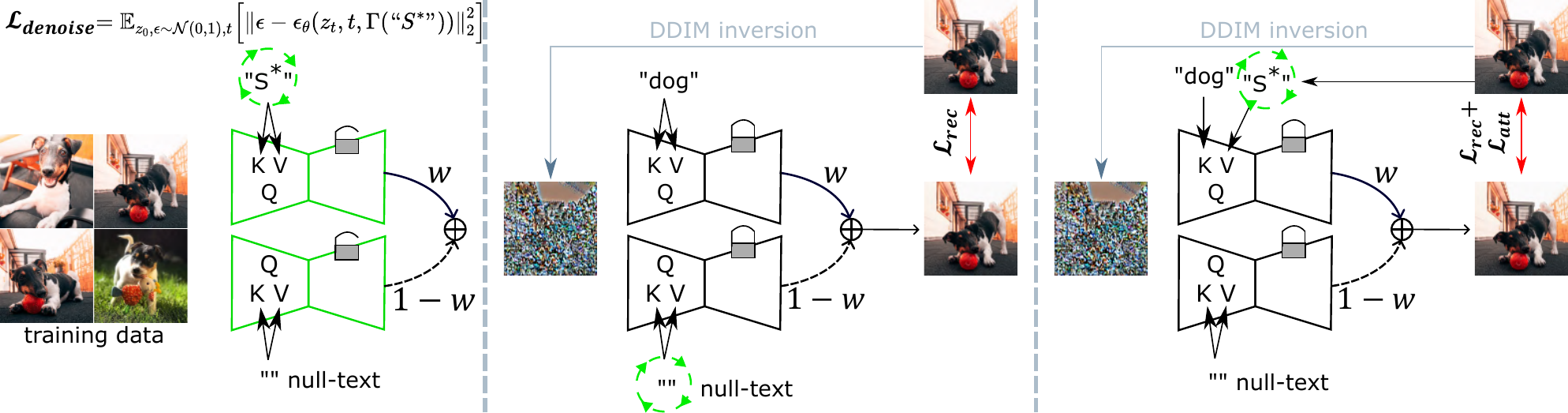}
        \caption{Three different optimization methods for inverting real image(s). (Left) Some works invert the image(s) into a new textual embedding “$S^{*}$” by finetuning the pretrained diffusion model
        or freezing the model 
        and applying a denoising loss $\mathcal{L}_{denoise}$. 
        Here, $w$ is the classifier-free guidance parameter. These methods require a few training images. 
        (Middle) Null-text inversion
        optimizes the null-text embedding with the reconstruction loss. (Right) Our Stylediffusion maps the real image to the input embedding of the \textit{value} of the cross-attention, enabling accurate style editing without causing significant structural changes.}
    \label{fig:survey}
\end{figure*}

\begin{figure*}[t]
	\centering
        \includegraphics[width=\linewidth]{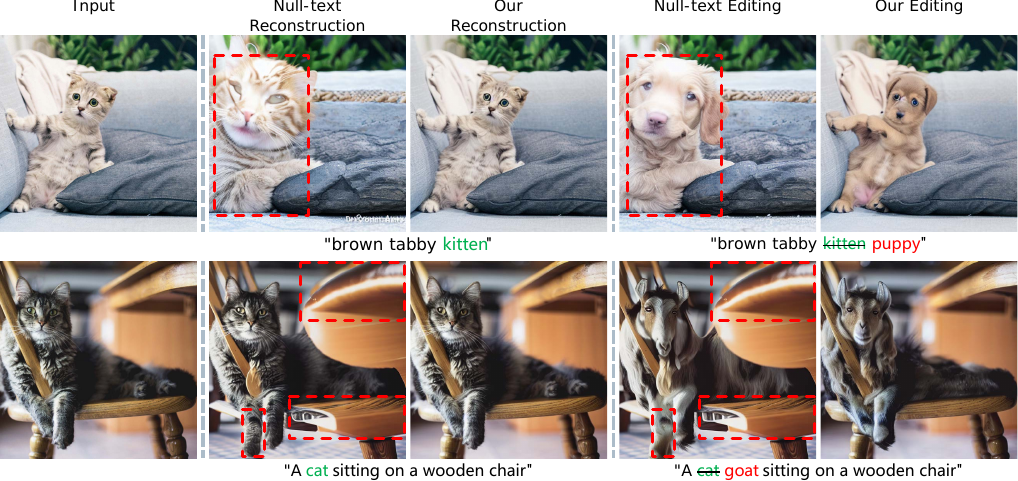}    
        \captionof{figure}{Our method takes as an input a real image (leftmost column) and an associated caption.  Here we demonstrate more accurate reconstruction and editing capabilities compared to Null-text
        . We manipulate the inverted image using the P2P editing technique.}
        \label{fig:teaser}
	\vspace{-3pt}
\end{figure*}
 
Text-based deep generative models have achieved extensive adoption in the field of image synthesis. Notably, GANs~\cite{patashnik2021styleclip,gal2021stylegan,kang2023gigagan,Sauer2023ICML}, diffusion models~\cite{saharia2022photorealistic,ramesh2022hierarchical,gafni2022make,rombach2021highresolution}, autoregressive models~\cite{yu2022scaling}, and their hybrid counterparts have been prominently utilized in this domain.

Diffusion models 
have made remarkable progress due to their exceptional realism and diversity. It has seen rapid applications in other domains such as video generation ~\cite{khachatryan2023text2video,wu2022tune,zhou2022magicvideo}, 3D generation ~\cite{poole2022dreamfusion,lin2023magic3d,wang2023prolificdreamer} and speech synthesis ~\cite{jeong2021diff,huang2022fastdiff,koizumi2022specgrad}.  In this work, we focus on Stable Diffusion (SD) models for real image editing. 
This capability is important for many real-world applications, including object replacement in images for publicity purposes (as well as personalized publicity), dataset enrichment where rare objects are added to datasets (e.g. wheelchairs in autonomous driving datasets), furniture replacement for interior design, etc.

\begin{figure*}[t]
    \centering
\includegraphics[width=\linewidth]{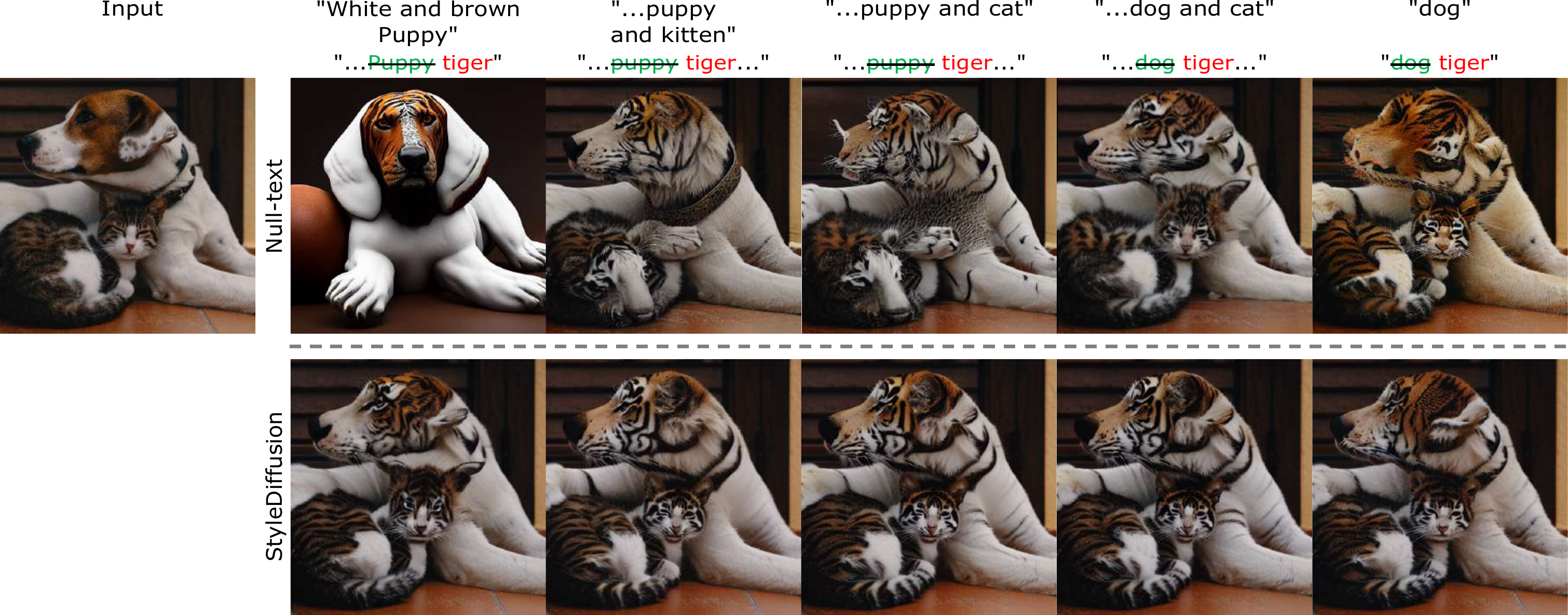}
        \caption{Null-text
        editing results of a real image with different prompts. A satisfactory result requires a carefully selected  prompt.}
    \label{fig:perfect_prompt}
\end{figure*}

\begin{figure}[t]
    \centering
\includegraphics[width=\columnwidth]{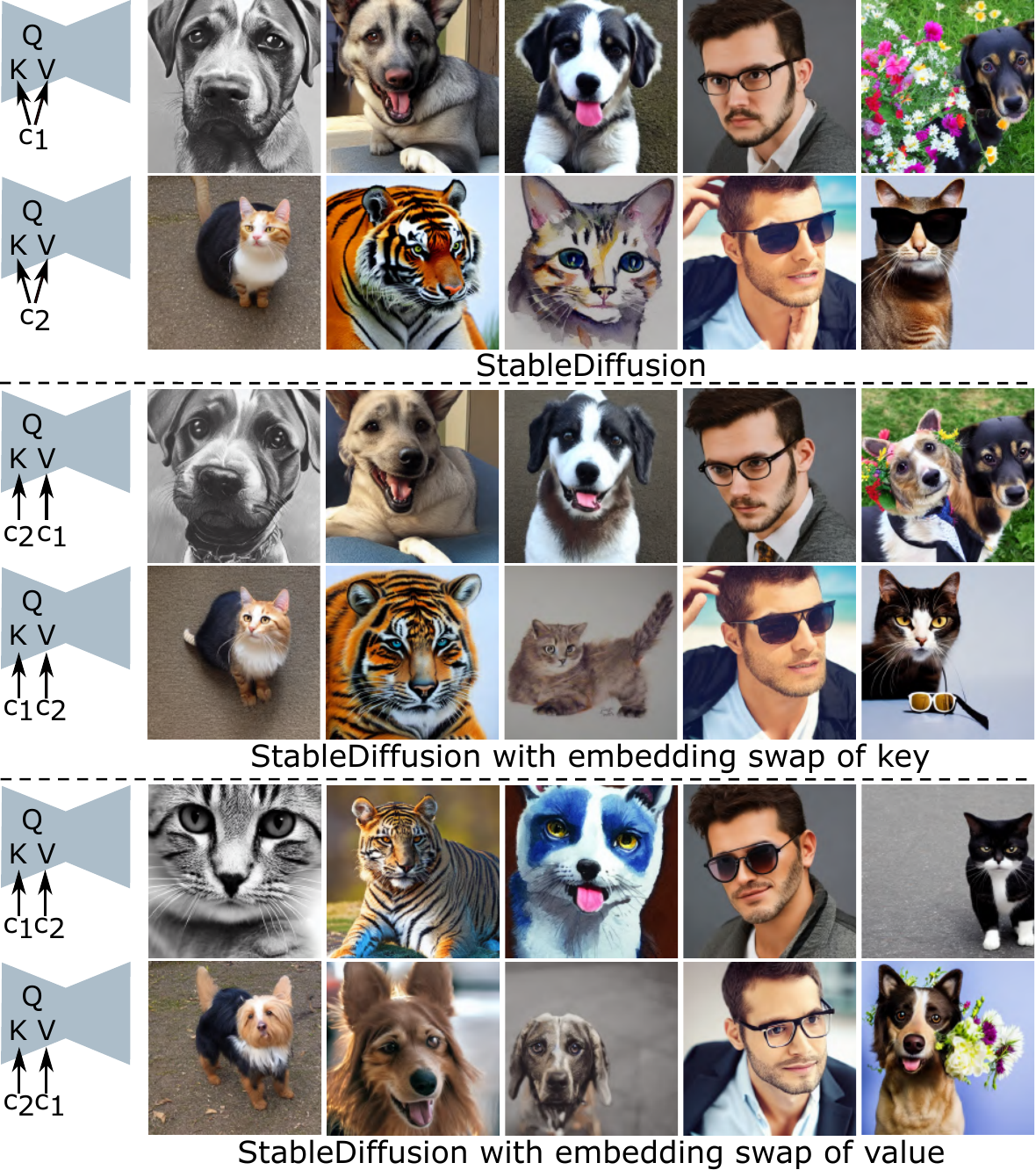}
        \caption{(Top) from second to last columns,  the prompts corresponding to both $\mathbf{c}_1$ and $\mathbf{c}_2$  are [\textit{“Dog”}, \textit{“Dog”}, \textit{“Dog”}, \textit{“A man in glasses”}, \textit{“A dog holding flowers”}] and [\textit{“Cat”}, \textit{“Tiger”}, \textit{“Watercolor drawing of a cat”}, “\textit{A man in sunglasses”}, \textit{“A cat wearing sunglasses”}], in turn. (Middle) we swap the input embedding of the \textit{key} and fix the one of the \textit{value}. This result shows that the generated images in both the third and fourth rows are similar to the ones in both the first and second rows, respectively. (Bottom) we fix the input embedding of the \textit{key} and swap the one of the \textit{value}. This experiment indicates that the value determines the object style (the ``what''). }
    \label{fig:k_v_exchange}
\end{figure}

Researchers basically perform real image editing with two steps: \textit{projection} and \textit{manipulation}. The former aims to either adapt the weights of the model~\cite{song2020denoising,liu2023accelerating,kim2022diffusionclip,xiao2023fastcomposer} or project given
images into the latent code or embedding space of SD models~\cite{mokady2022null,gal2022image,avrahami2023break,han2023highly}.
The latter aims to edit the latent code or embedding to further manipulate real images~\cite{Kawar2022ImagicTR,meng2021sdedit,cao2023masactrl,zhang2023adding,couairon2022diffedit}. In the first \emph{projection} step, some works finetune the whole model~\cite{Kawar2022ImagicTR, valevski2022unitune, ruiz2022dreambooth,kim2022diffusionclip,gal2023encoder,xiao2023fastcomposer} or  partial weights of the pretrained model~\cite{kumari2022multi,xie2023difffit}.  Yet,  finetuning either the entire or part of the generative model with only a few real images suffers from both the cumbersome tuning of the model's weights and catastrophic forgetting~\cite{kumari2022multi,wu2018memory,xie2023difffit}.  
Other works on \emph{projection} attempt to  learn a new embedding vector which represents given real images (keeping the SD model frozen)~\cite{ho2021classifier,gal2022image,avrahami2023break,han2023highly,tewel2023key,zhang2023prospect,dong2023prompt}. They focus on optimizing conditional or unconditional inputs of the cross-attention layers of the classifier-free diffusion model~\cite{ho2021classifier}. 
Textual Inversion~\cite{gal2022image} uses the denoising loss  to optimize the  textual embedding of the conditional branch given a few content-similar images. 
Null-text optimization~\cite{mokady2022null} firstly inverts the real image into a series of timestep-related latent codes, then leverages a reconstruction loss to learn the null-text embedding of the unconditional branch (see Fig.~\ref{fig:survey}(middle)). 
However, these methods suffer from the following issues. \textbf{Firstly}, they lead to unsatisfactory results for selected regions, and unexpected changes in non-selected regions, both during reconstruction and editing (see the Null-text results in Fig.~\ref{fig:teaser}; the structure-dist ($\downarrow$) for Null-text Editing and editing by our method are 0.046 and 0.012 in the first row, and  0.019 and 0.054 in the second row). \textbf{Secondly}, they require a user to provide an accurate text prompt that describes every visual object,  and the relationships between them in the input image (see Fig.~\ref{fig:perfect_prompt}).
\textbf{Finally}, Pix2pix-zero~\cite{parmar2023zero} requires the textual embedding directions (e.g, cat $\rightarrow$ dog in Fig.~\ref{fig:cat2dog}(up, the fourth column)) with thousands of sentences with GPT-3~\cite{brown2020language} before editing, which lacks scalability and flexibility.

To overcome the above-mentioned challenges, we analyze the role of the attention mechanism (and specifically the roles of keys, queries and values) in the diffusion process. This leads to the observation that the key dominates the output image structure (the ``where'')~\cite{hertz2022prompt},  whereas the value determines the object style (the ``what'').  
We perform an effective experiment to demonstrate that the value determines the object style (the ``what''). As shown in Fig.~\ref{fig:k_v_exchange}(top),  we generate two sets of images with prompt embeddings $\mathbf{c}_1$ and $\mathbf{c}_2$.
We use two different embeddings for the input of both key and value in the same attention layer. When swapping the input of the keys and fixing the input of the values, we observe that the content of generated images are similar,  see Fig.~\ref{fig:k_v_exchange}(middle). For example, The images of Fig.~\ref{fig:k_v_exchange}(middle) are similar to the ones of  Fig.~\ref{fig:k_v_exchange}(top).  When exchanging the input of the values and fixing the one of the keys, we find that the content swaps while preserving much of the structure,  see Fig.~\ref{fig:k_v_exchange}(bottom). For example, the images of the last row of Fig.~\ref{fig:k_v_exchange}(bottom) have similar semantic information  with the ones of the first row of Fig.~\ref{fig:k_v_exchange}(top).  It should be noted that their latent codes are all shared\footnote{The first row of top, middle, and bottom shares a common latent code, and the second row also shares a common latent code.}, so the structure of the results in Fig.~\ref{fig:k_v_exchange}(middle) does not change significantly.
This experiment indicates that the value determines the object's style (the ``what''). 

Therefore, to improve the projection of a real image, we introduce $\textbf{Stylediffusion}$ which maps a real image to the input embedding for the value computation (we refer to this embedding as the $\textbf{prompt-embedding}$),  which enables us to obtain accurate style editing without invoking significant structural changes. 
We propose to map the real image to the input of the \textit{value} linear layer in the cross-attention layers~\cite{bahdanau2014neural,NIPS2017_3f5ee243} providing freedom to edit effectively the real image in the manipulation step.

We take the given textual embedding as  the input of the \textit{key} linear layer, which is frozen (see Fig.~\ref{fig:survey}(right)). Using frozen embedding contributes to preserving the well-learned attention map from DDIM inversion, which guarantees the initial editability of the inverted image. 
We observe that the system often outputs unsatisfactory reconstruction results (greatly adjusting the input image structure) (see Fig.~\ref{fig:teaser}(first row, second column) and Fig.~\ref{fig:attnloss}(first row, second column)) due to locally less accurate attention maps (see Fig.~\ref{fig:attnloss}(first row, and third column)). 
Hence, to further improve our method, we propose an attention regularization to obtain more precise reconstruction and editing capabilities.

\begin{figure}[t]
    \centering
    \includegraphics[width=\columnwidth]{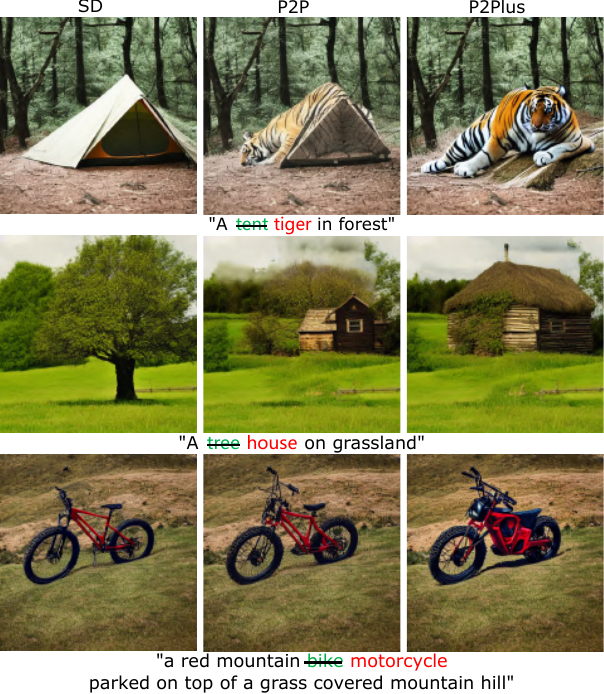}
        \caption{
Comparison of both P2P and P2Plus. P2P fails when the desired editing changes require large structural changes (e.g., tent, tree, and bike). By reducing the dependence on the source prompt (by also involving the unconditional branch), P2Plus can better handle such cases. 
The Clipscore ($\uparrow$) of P2P and P2Plus are 84.2\% and 87.2\% in the first row, 77.6\% and 78.4\% in the second row, and 70.8\% and 78.6\% in the third row.
        }
    \label{fig:p2plus}
\end{figure}

For the second manipulation step, researchers propose a series of outstanding techniques~\cite{Kawar2022ImagicTR,meng2021sdedit,cao2023masactrl,zhang2023adding,couairon2022diffedit,mou2023dragondiffusion,jia2023taming,zhang2023continuous,qiu2023controlling,levin2023differential}. Among them, P2P~\cite{hertz2022prompt} is one of the most widely used image editing methods. 
However, P2P only operates on the conditional branch, and ignores the unconditional branch. This leads to less accurate editing capabilities for some cases, especially where the structural changes before and after editing are relatively large (e.g., “...tree...”$\rightarrow$“...house...” in Fig.~\ref{fig:p2plus}).  To address this problem, we need to reduce the dependence of the structure on the source prompt and provide more freedom to generate the structure following the target prompt. Since the unconditional branch allows us to edit out  concepts~\cite{armandpour2023re,Tumanyan_2023_CVPR}. Thus, we propose to further perform the self-attention map exchange in  the unconditional branch based on P2P (referred to as \textit{P2Plus}), as well as in the conditional branch like P2P~\cite{hertz2022prompt}. This technique enables us to obtain more accurate editing capabilities (see Fig.~\ref{fig:p2plus}(third column)). We build our method on SD models~\cite{rombach2021highresolution} and experiment on a variety of images and several ways of prompt editing.

Our work thus makes the following contributions:
\begin{itemize}[leftmargin=*]
    \item State-of-the-art methods (e.g., Null-text inversion) struggle with unsatisfactory reconstruction and editing. To precisely project a real image, we introduce \textbf{StyleDiffusion}. We use a simple mapping network to map a real image to the input embedding for \textit{value} computation.
    \item We propose an attention regularization method to enhance the precision of attention maps, resulting in more accurate reconstructions and improved editing capabilities.
    \item We propose the P2Plus technique, which enables us to obtain more powerful editing capabilities, especially when the source object is unrelated to the target object.    
    This approach addresses the limitations of P2P, which fails to function effectively in such cases.
    \item Through extensive experiments, we demonstrate the effectiveness of our method for accurately reconstructing and editing images.
\end{itemize}

\section{Related work}

\subsection{Transfer learning for diffusion models}
A series of recent works has investigated knowledge transfer on diffusion models~\cite{song2020denoising,liu2023accelerating,mokady2022null,gal2022image,avrahami2023break,kwon2022diffusion,kim2022diffusionclip,gal2023encoder,xiao2023fastcomposer,han2023highly,tewel2023key,kumari2023multi,zhang2023prospect,dong2023prompt} with one or a few images. 
Recent work~\cite{Kawar2022ImagicTR,meng2021sdedit,cao2023masactrl,zhang2023adding,couairon2022diffedit,mou2023dragondiffusion,jia2023taming,zhang2023continuous,qiu2023controlling,levin2023differential,chen2023training}  either finetune the pretrained model or invert the image in the latent space of the  pretrained model. 
Dreambooth~\cite{ruiz2022dreambooth} shows that training a diffusion model on a small data set (of $3\sim5$ images) largely benefits from a pre-trained diffusion model, preserving the textual editing capability. Similarly,  Imagic~\cite{Kawar2022ImagicTR} and UniTune~\cite{valevski2022unitune} rely on the interpolation weights or the classifier-free guidance at the inference stage, except when finetuning the diffusion model during training. Kumari et al.~\cite{kumari2022multi} study only updating part of the parameters of the pre-trained model, namely the \textit{key} and \textit{value} mapping from text to latent features in the cross-attention layers.  
However, updating the diffusion model unavoidably loses the text editing capability of the pre-trained diffusion model. In this paper, we focus on real image editing with a frozen text-guilded diffusion model.

\subsection{GAN inversion}
Image inversion aims to project a given real image into the latent space,  allowing users to further manipulate the image. There exist several approaches~\cite{creswell2018inverting,goetschalckx2019ganalyze,jahanian2019steerability,lipton2017precise,  xia2021gan,yeh2017semantic}  which focus on image manipulation based on pre-trained GANs, following literately optimization of the latent representation to restructure the target image.  Given a target semantic attribute, they aim to manipulate the output image 
of a pretrained GAN. Several other methods~\cite{abdal2019image2stylegan,zhu2020domain} reverse a given image into the input latent space of a pretrained GAN (e.g., StyleGAN), 
and restructure the target image by optimization of the latent representation.  They mainly consist of  fixing the generator~\cite{abdal2019image2stylegan,abdal2020image2stylegan++, richardson2020encoding,tov2021designing} or updating the generator~\cite{alaluf2021hyperstyle,roich2021pivotal}.

\subsection{Diffusion model inversion}
Diffusion-based inversion can be performed naively by optimizing the latent representation.  \cite{dhariwal2021diffusion} show that a given real image can be reconstructed by DDIM sampling ~\cite{song2020denoising}. DDIM provides a good starting point to synthesize a given real image.  Several works~\cite{avrahami2022blendedlatent, avrahami2022blended,nichol2021glide} assume that the user provides a mask to restrict the region in which the changes are applied, achieving both meaningful editing and background preservation. P2P~\cite{hertz2022prompt} proposes a mask-free editing method. However, it leads to unexpected results when editing the real image~\cite{mokady2022null}.  Recent work investigates the text embedding of the conditional input~\cite{gal2022image},  or the null-text optimization of the unconditional input (i.e., Null-text inversion~\cite{ mokady2022null}).  Although having the editing capability by combining the new prompts,  
they suffer from the following challenges
(i) they lead to unsatisfying results for the selected regions, and unexpected changes in non-selected regions, and (ii) they require careful text prompt editing where the prompt should include all visual objects in the input image.

Concurrent work~\cite{parmar2023zero} proposes pix2pix-zero, also aiming to provide more accurate editing capabilities of the real image. However, it firstly needs to compute the textual embedding direction in advance using thousand sentences.

\begin{figure*}[ht!]
    \centering
\includegraphics[width=\textwidth]{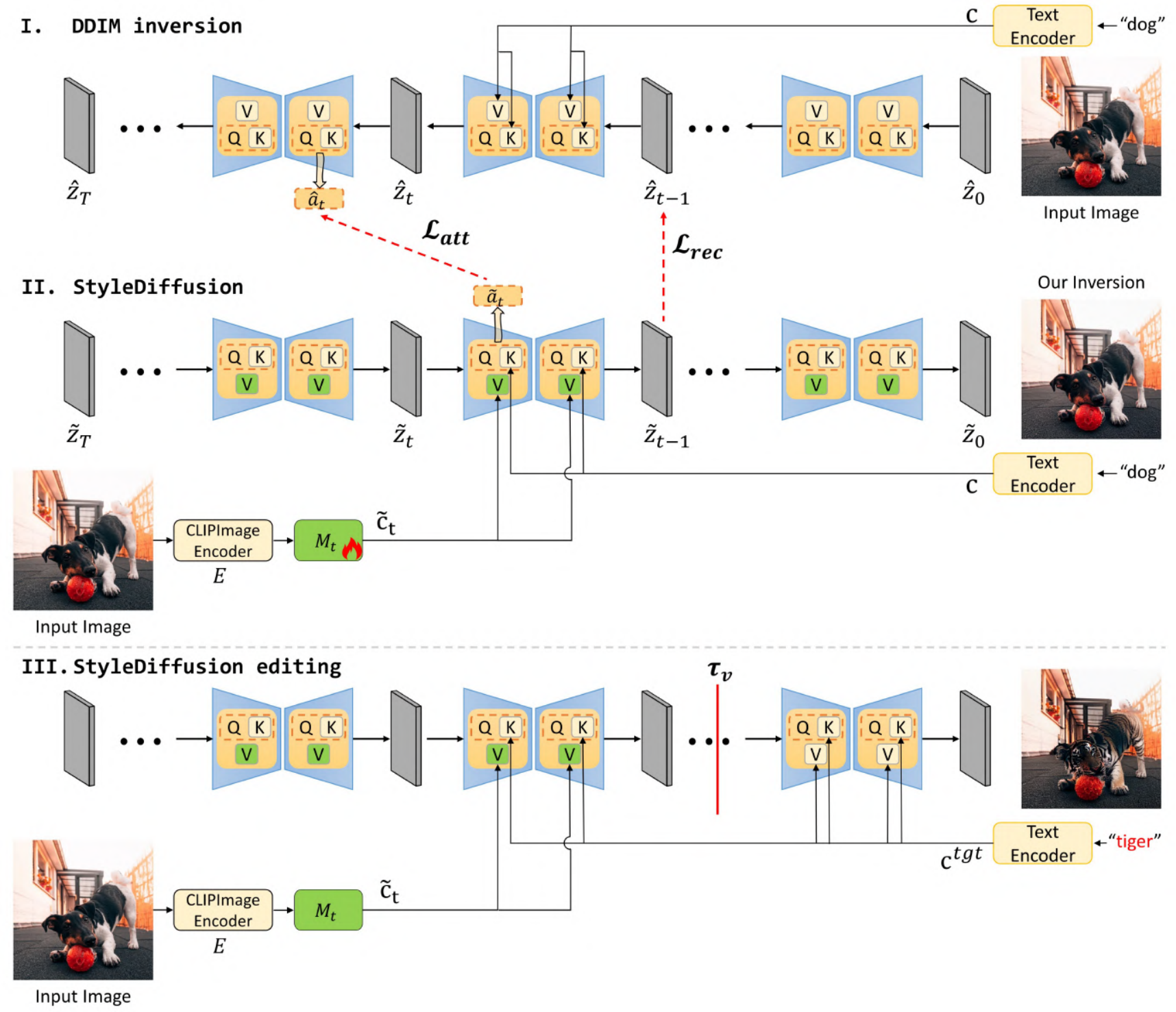}
        \caption{Overview of the proposed method. (I) DDIM inversion: the diffusion process is performed to generate the latent representations: ${(\mathbf{\hat{z}}_t, \mathbf{\hat{a}}_t)}, (t = 1,...,T)$, where $\mathbf{\hat{z}}_0 = \mathbf{z}_0$, which is the extracted feature of the input real image $\mathbf{x}$. $\mathbf{c}$ is the textual embedding extracted by a Clip-text Encoder with a given prompt $\mathbf{p}^{src}$.    (II) Stylediffusion: we take the input image $\mathbf{x}$ as input, and extract the prompt-embedding $\mathbf{\widetilde{c}}_{t} = M_{t}\left ( E(\mathbf{\mathbf{x}})\right )$, which is taken as the input value matrix $\mathbf{v}$ of the linear layer $\Psi_V$. The input of  the linear layer $\Psi_K$ is the given textual embedding $\mathbf{c}$.  We get both the latent code $\mathbf{\widetilde{z}_{t-1}}$ and the attention map $\mathbf{\widetilde{a}}_t$, which are aligned with both the latent code $\mathbf{\hat{z}_{t-1}}$ and the attention map $\mathbf{\hat{a}_{t}}$, respectively.  Note that $\mathbf{\widetilde{z}}_T = \mathbf{\hat{z}}_T$.     (III) StyleDiffusion editing: for timesteps from T to $\tau_v$, the input of the linear network $\Psi_v$ comes from the learned textual embedding $\mathbf{\widetilde {c}_{t}}$ produced by the trained $M_t$. From $\tau_{v}-1$ to 1 the corresponding input comes from the prompt-embedding $\mathbf{c}^{tgt}$ of the target prompt. We use P2Plus to perform the attention exchange.}
    \label{fig:framework}
\end{figure*}

\section{Method}

\subsection{Background}
DDIM inversion proposes an inversion scheme for unconditional diffusion models. However, this method fails when applied to text-guided diffusion models. This was observed by Mokady et al.~\cite{mokady2022null}, who propose Null-text inversion to address this problem. However, their methods has some drawbacks:  (1) unsatisfying results for selected regions and unexpected changes in non-selected regions, and (2) they require careful text prompt editing where the prompt should include all visual objects in the input image.

Therefore, our goal is to obtain a more accurate editing capability based on an accurate reconstruction of the real image $\mathbf{x}$ guided by the source prompt $\mathbf{p}^{src}$.      
Our method, called StyleDiffusion, is based on  the observation that the \emph{keys} of the cross-attention layer dominate the output image structure (the ``where''),  whereas the \emph{values} determine the object style (the ``what''). After faithfully projecting the real image, we propose P2Plus, which is an improved version of P2P~\cite{hertz2022prompt}. 

Next, we  introduce SD models in Sec.~\ref{subsec:SD_models}, followed by the proposed StyleDiffusion in Sec.~\ref{subsec:stylediffusion} and P2Plus in Sec.~\ref{subsec:p2plus}. A general overview is provided in Fig.~\ref{fig:framework}.

\subsection{Preliminary: Diffusion Model}\label{subsec:SD_models}
\subsubsection{Basis}
Generally, diffusion models optimize a UNet-based denoiser  network $\epsilon_\theta$  to predict Gaussian noise $\epsilon$,  following the objective:
\begin{equation}
\min_\theta E_{\mathbf{z}_0,\epsilon \sim N(0,I),t\sim [1,T]} \left \| \epsilon-\epsilon_\theta(\mathbf{z}_t,t,\mathbf{c}) \right \|_{2}^{2}, 
\end{equation}
where $z_t$ is a noise sample according to timestep $t\sim [1,T]$, and $T$ is the number of the timesteps.  The encoded text embedding $\mathbf{c}$ is extracted by a Clip-text Encoder $\Gamma$ with given prompt $\mathbf{p}^{src}$: $\mathbf{c} = \Gamma(\mathbf{p}^{src})$. 
In this paper, we  build on SD models~\cite{rombach2021highresolution}.  These first train both encoder  and decoder. Then the diffusion process is performed in the latent space. Here the encoder maps the image $\mathbf{x}$ into the latent representation  $\mathbf{z_0}$, and the decoder aims to reverse the  latent representation  $\mathbf{z_0}$ into the image.  The sampling process is given by:
\begin{eqnarray}
\mathbf{z}_{t-1} = \sqrt{\frac{\alpha_{t-1}}{\alpha_t}}\mathbf{z}_t 
+\nonumber\\ \sqrt{\alpha_{t-1}}
\left(\sqrt{\frac{1}{\alpha_{t-1}}-1}-\sqrt{\frac{1}{\alpha_t}-1}\right) \epsilon_\theta(\mathbf{z}_t,t,\mathbf{c}),
\label{eq:ddim_sampling}
\end{eqnarray}
 where  $\alpha_t$ is a scalar function.

\subsubsection{DDIM inversion}
For real-image editing with a pretrained diffusion model, a given real image is to be reconstructed by finding its initial noise. 
 We use the deterministic DDIM model to perform image inversion.  
 This process is given by:
\begin{eqnarray}
\label{eq:ddim_inversion}
\mathbf{z}_{t+1} = \sqrt{\frac{\alpha_{t+1}}{\alpha_t}}\mathbf{z}_t
+ \nonumber\\ \sqrt{\alpha_{t+1}}
\left(\sqrt{\frac{1}{\alpha_{t+1}}-1}-\sqrt{\frac{1}{\alpha_t}-1}\right) \epsilon_\theta(\mathbf{z}_t,t,\mathbf{c}).
\end{eqnarray}
 DDIM inversion synthesizes the latent noise that produces an approximation of the input image when fed to the diffusion process. While the reconstruction based on DDIM is not sufficiently accurate,  it still provides a good starting point for training, enabling us to efficiently achieve high-fidelity  inversion~\cite{hertz2022prompt}. We use  the intermediate results of DDIM inversion to train our model, similarly as~\cite{couairon2022diffedit,mokady2022null}.

\subsubsection{Cross-attention}
SD models achieve text-driven image generation by feeding a prompt into the cross-attention layer.  Given both the text embedding $\mathbf{c}$ and  the image feature representation $\mathbf{f}$,  
we are able to produce the key matrix $\mathbf{k}= \Psi_K(\mathbf{c})$,  the value matrix $\mathbf{v}= \Psi_V(\mathbf{c})$ and the query matrix $\mathbf{q}= \Psi_Q(\mathbf{f})$,  via the  linear networks:  $\Psi_K, \Psi_V,\Psi_Q$. The attention maps are then  computed with:
\begin{equation}
\begin{aligned}\label{eq:atten_softmax}
\mathbf{a}= \text{Softmax}({\mathbf{q}\mathbf{k}^{T}}/{\sqrt{\mathbf{d}}}),
\end{aligned}
\end{equation}
where $\mathbf{d}$ is the projection dimension of the keys and queries.  Finally, the cross-attention output is $\mathbf{\hat{f}} = \mathbf{a}\mathbf{v}$, which is then taken as input in the following  convolution layers. 

Intuitively, P2P~\cite{hertz2022prompt} performs prompt-to-prompt image editing  with cross attention control. P2P is based on the idea that the attention maps largely control where the image is drawn, and the values decide what is drawn (mainly defining the style). 
Improving the accuracy of the attention maps leads to more powerful editing capabilities~\cite{mokady2022null}.
We experimentally observe that DDIM inversion generates satisfying attention maps (e.g., Fig.~\ref{fig:attnloss}(second row, first column)), 
and provides a good starting point for the optimization. Next, we investigate the attention maps to guide the image inversion.

\subsection{StyleDiffusion}
\label{subsec:stylediffusion}

\begin{algorithm}[t]
\SetAlgoLined
\textbf{Require:} the features of the training image and the prompt embeddings: $\{\mathbf{z}_0,  \mathbf{c} \}$. $K_t=e^{-t}*K$, where $K=100$, the starting number of inner iterations.
$K_t$ is the number of training iteration for each timestep t.
The mapping network $\{M_t\}, (t=1,...,T)$ with initialization parameters $\omega$.\\
\textbf{Temporary results:} With guidance scale $w=1$ for the classifier-free diffusion model, we use DDIM inversion to produce $\{ \mathbf{\hat{z}}_j, \mathbf{\hat{a}}_j\}, (j=1,...,T)$.

\textbf{Output:} 
Mapping network $\{M_t\} (t=1,...,T)$.\\
 \vspace{1mm} \hrule \vspace{1mm}
 Set guidance scale $w=7.5$; \\
 Initializing $ \mathbf{\widetilde{z}}_T \leftarrow \mathbf{\hat{z}}_T$; \\
 \For{$t=T,T-1,\ldots,1$}{
    \For{$k=0,\ldots,K_t-1$}{
        $ {\mathbf{a}_{t}, \mathbf{z}_{t-1}} \leftarrow \mathbf{\widetilde{z}}_t$;(Eqs.~\ref{eq:atten_softmax} and \ref{eq:pred_noise})\\
        $\omega  \leftarrow  \omega  - \eta \nabla_{\omega }\mathcal{L}$ ;(Eq.~\ref{eq:full_loss})
    }
 Synthesizing $\mathbf{\widetilde{z}}_{t-1}$;(Eq.~\ref{eq:pred_noise})
}
\textbf{Return} Mapping network $\{M_t\}, (t=1,...,T)$
\caption{Our algorithm}\label{alg:alg_ours}
\end{algorithm}

\subsubsection{Method overview}
As shown in Fig.~\ref{fig:framework}(I),  given a pair of a real image $\mathbf{x}$ and a corresponding prompt $\mathbf{p}^{src}$ (e.g., "dog"),  we  perform DDIM inversion~\cite{dhariwal2021diffusion,song2020denoising} to synthesize a series of latent noises $\{\mathbf{\hat{z}}_t\}$ and attention maps $\{\mathbf{\hat{a}}_t\} (t = 1, ..., T)$, where $\mathbf{\hat{z}}_0 = \mathbf{z}_0$, which is the extracted  latent code of the input image $\mathbf{x}$~\footnote{Note  when generating the attention map $\mathbf{\hat{a}}_T$ in the last timestep $t=T$, we throw out the synthesized latent code $\mathbf{\hat{z}}_{T+1}$. }.
Fig.~\ref{fig:framework}(II)   shows that our method reconstructs the latent noise $\mathbf{\hat{z}}_t$ in the order of the diffusion process $T \rightarrow 0$, where $\mathbf{\widetilde{z}}_T=\mathbf{\hat{z}}_T$. Our framework consists of three networks: a frozen ClipImageEncoder $E$,   a learnable mapping network $M_{t}$ and a denoiser network $\epsilon_\theta$.  For a specific timestep $t$, the  ClipImageEncoder $E$ takes the input image $\mathbf{x}$ as an input. The output $E(\mathbf{x})$ is fed into the mapping network $M_{t}$, producing the prompt-embedding $\mathbf{\widetilde {c}_{t}} = M_{t}\left ( E(\mathbf{x}) \right )$, which is fed into the value network $\Psi_V$ of the cross-attention layers. The  input of the linear layer $\Psi_K$ is the given textual embedding $\mathbf{c}$. We generate both the latent code $\mathbf{\widetilde{z}_{t-1}}$ and the attention map $\mathbf{\widetilde{a}}_t$. 
Our full algorithm is presented in algorithm~\ref{alg:alg_ours}.

The full loss function consists of two losses:   \textit{reconstruction loss} and \textit{attention loss}, which   guarantee
that both  the denoised latent code $\mathbf{\widetilde{z}_{t-1}}$ and  the  corresponding attention map $\mathbf{\widetilde{a}_{t}}$  at inference time are close to the ones: $\mathbf{\hat{z}_{t-1}}$ and  $\mathbf{\hat{a}_{t}}$ from DDIM inversion, respectively. 

\subsubsection{Reconstruction loss}
Since the noise representations ($\{\mathbf{\hat{z}}_1, \cdots \mathbf{\hat{z}}_T\}$) provide an initial trajectory which is close to the real image, we train the network $M_{t}$
to generate the prompt embedding $\mathbf{\widetilde {c}_{t}} = M_{t}\left ( E(\mathbf{x}) \right )$, which is the input of the value network. We optimize the $M_{t}$ in such a manner, that the output latent code ($\mathbf{\widetilde{z}}_t$)  is close to the noise representations ($\mathbf{\hat{z}}_t$). The objective is:

\begin{equation}\label{eq:recon}
\resizebox{0.5\hsize}{!}{$%
 \mathcal{L}_\mathrm{rec} = \min_{M_{t}}  \left \|\mathbf{\hat{z}}_{t-1}-\mathbf{\widetilde{z}}_{t-1}\right \|^2
 $,\vspace{-10mm}}
\end{equation}
\begin{eqnarray}\label{eq:pred_noise}
 \mathbf{\widetilde{z}}_{t-1} = \sqrt{\frac{\alpha_{t-1}}{\alpha_t}}\widetilde{\mathbf{z}}_t+\nonumber\\
 \sqrt{\alpha_{t-1}}
 \left(\sqrt{\frac{1}{\alpha_{t-1}}-1}-\sqrt{\frac{1}{\alpha_t}-1}\right) \epsilon_\theta(\widetilde{\mathbf{z}}_t,t, \mathbf{c}, \mathbf{{c}_{t}}).
\end{eqnarray}

\begin{figure}[t]
    \centering
    \includegraphics[width=\columnwidth]{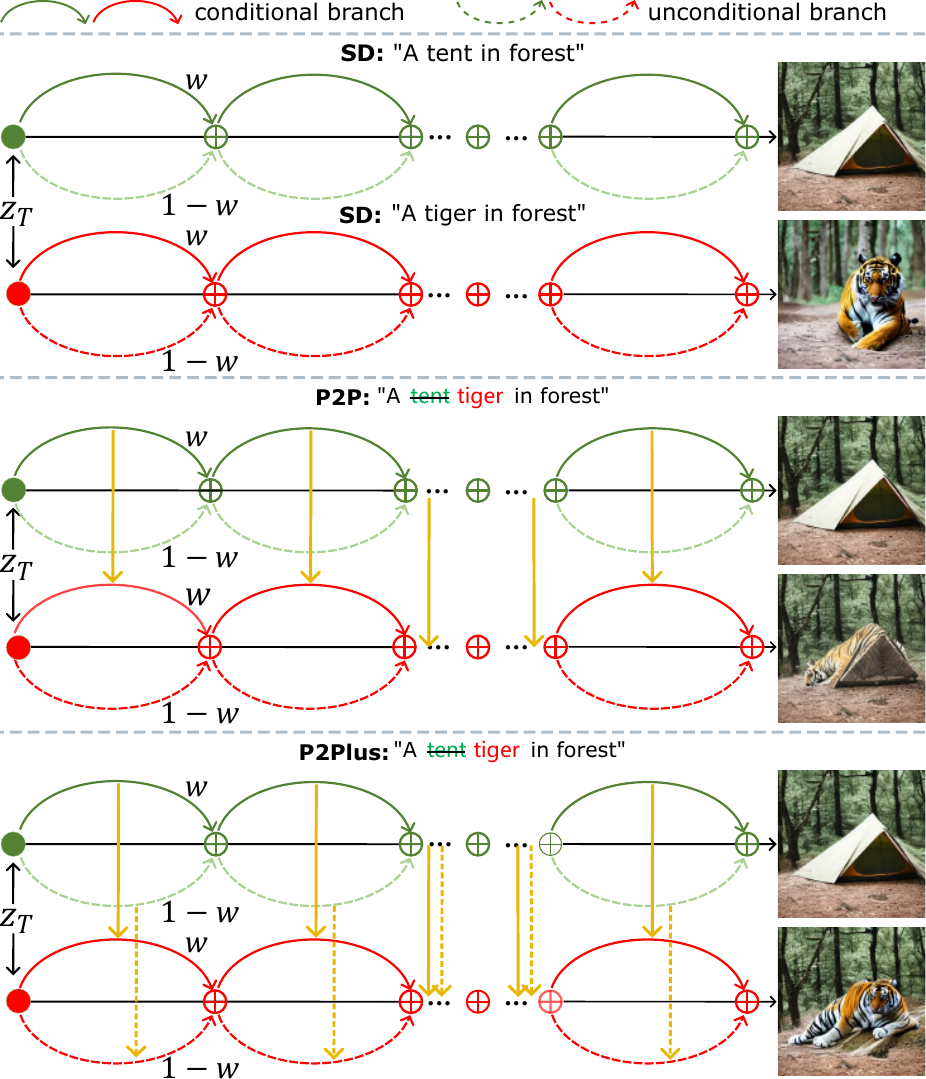}
        \caption{(Top) Given the same latent code $\mathbf{z}_{T}$ and the classifier-free guidance parameter $w$, we generate two images with two prompts. The arrows indicate the diffusion process from $T$ to $0$. Here \textcolor{green}{green} indicates generation using the source prompt: ``A tent in forest'' and  \textcolor{red}{red} indicates generation using the target prompt: ``A tiger in forest''. (Middle) P2P copies both the self-attention maps and the cross-attention maps of the conditional branch of the SD models from the source prompt into the corresponding self-attention and cross-attention maps of the conditional branch of the SD models with target prompt. This process is indicated by  \textcolor{yellow}{yellow} arrows.  (Bottom) P2Plus additionally replaces the self-attention map of the unconditional branch of SD models, in addition to the ones of the conditional branch. The dashed \textcolor{yellow}{yellow} arrows indicate this technique. }
    \label{fig:p2plus_pipline}
\end{figure}

\subsubsection{Attention loss}
It is known that a more accurate attention map is positively correlated to the editing capability~\cite{mokady2022null}. 
The attention map, which is synthesized during the DDIM inversion, provides a good starting point. Thus, we introduce attention regularization 
when optimizing the mapping network $M_{t}$ to further improve its quality. The objective is the following:
\begin{equation}\label{eq:atten}
 \resizebox{0.5\hsize}{!}{$%
 \mathcal{L}_\mathrm{att} = \min_{M_{t}}  \left \|\mathbf{\hat{a}}_{t}-\mathbf{\widetilde{a}}_{t}\right \|^2
 $,\vspace{-10mm}}
\end{equation}
where  $\mathbf{\hat{a}}_{t}$ and  $\mathbf{\widetilde{a}}_{t}$ can be obtained with Eq.~\ref{eq:atten_softmax}. 

\subsubsection{Full objective}
The full objective function of our model is:
\begin{equation}
\label{eq:full_loss}
\begin{aligned}
\mathcal{L} =   \mathcal{L}_\mathrm{rec} +\mathcal{L}_\mathrm{att}.
\end{aligned}
\end{equation}
In conclusion, in this section, we have proposed an alternative solution to the inversion of text-guided diffusion models which aims to improve upon existing solutions by providing more accurate editing capabilities, without requiring careful prompt engineering. 

\subsection{P2Plus}
\label{subsec:p2plus}

\begin{figure*}[t]
   \centering
\includegraphics[width=\textwidth]{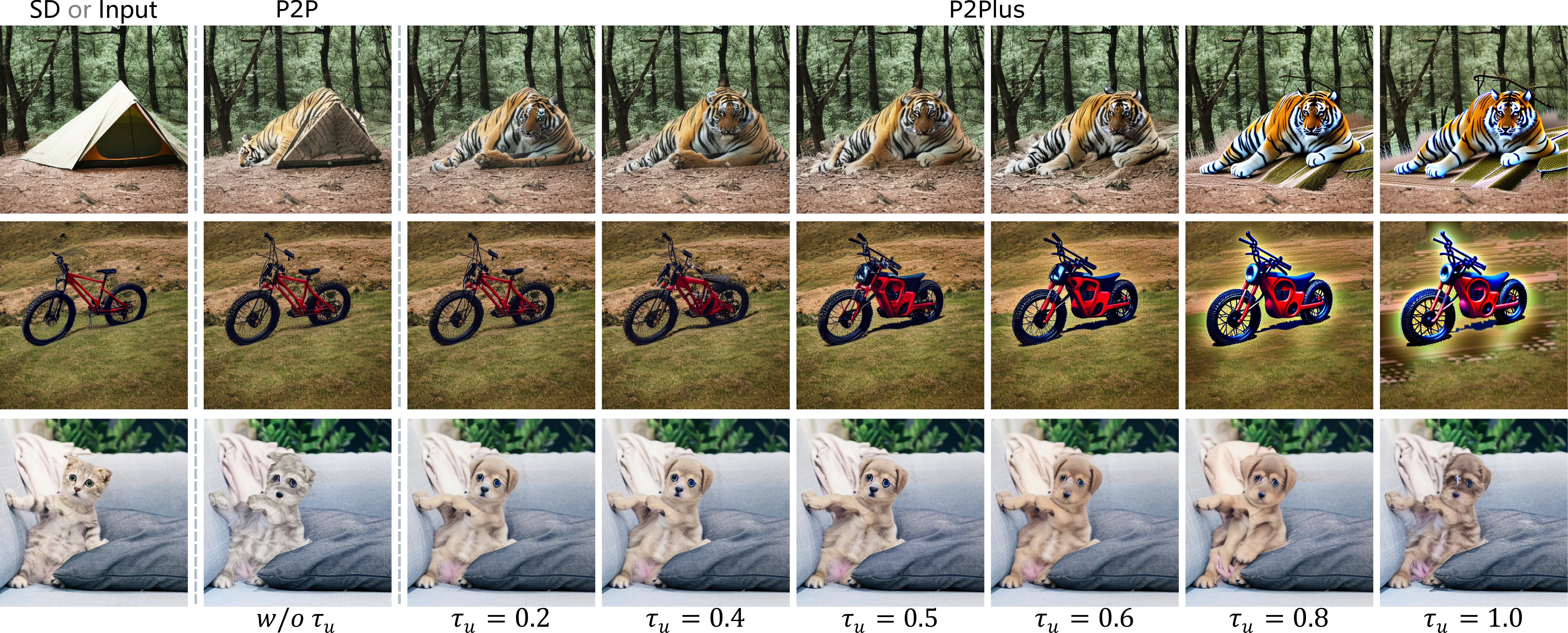}
       \caption{\emph{P2P} (i.e., second column: w/o $\tau_u$) has not succeeded in replacing the tent by the tiger.
       Adding the injection parameter $\tau_u$ can help to edit successfully, especially if $\tau_u=0.5$.
       We also use the classifier-free guidance parameter $w=7.5$ like SD, when the weight $1-w$ of the unconditional branch is negative
       , which can gradually weaken the influence of the ``tent'' in the unconditional branch as $\tau_u$ increases from 0.2 to 1.0 (third to eighth columns). 
       The source images in the first and second rows are from SD, while the source image in the third row is from the input real image (first column). Editing from the first to the third row are ``...tent...'' $\rightarrow$ ``...tiger...'', ``...bike...'' $\rightarrow$ ``...motorcycle...'', and ``...kitten...'' $\rightarrow$ ``...puppy...''.
       }
   \label{fig:uncondselfattn}
\end{figure*}

\begin{figure*}
    \centering
    \includegraphics[width=0.94\linewidth]{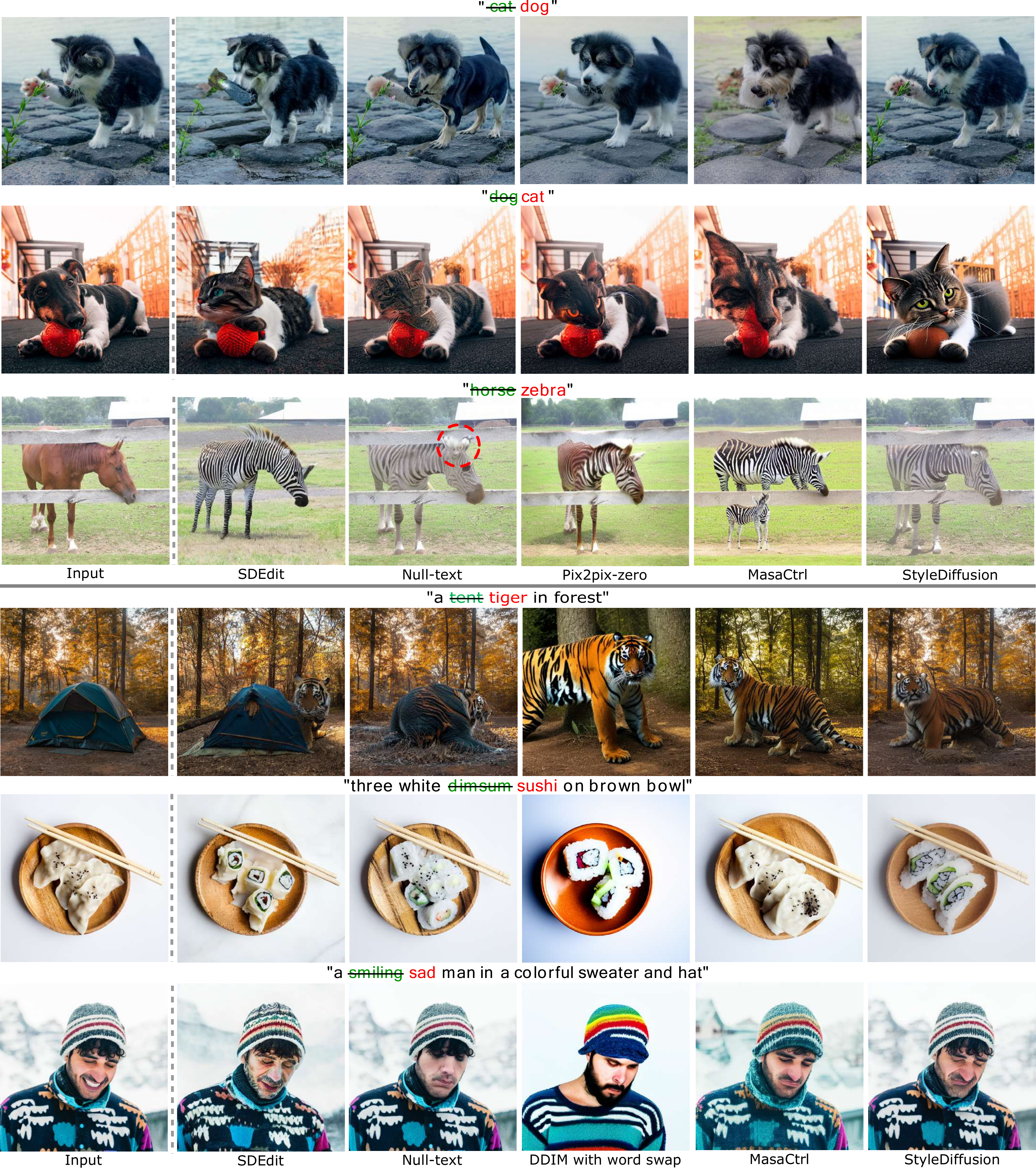}
    \caption{Comparisons with different baselines for real images.  
        Our method achieves realistic editing of both style and structured objects, while preserving the structure of the input image (last column).  }
    \label{fig:cat2dog}
\end{figure*}

Having inverted the text-guided diffusion model, we can now perform prompt-based image editing (see Figs.~\ref{fig:teaser} and \ref{fig:cat2dog}). We here outline our approach, 
which improves on the popular P2P

P2P performs the replacement of both the cross-attention map and the self-attention map of the conditional branch,  aiming to maintain the structure of the source prompt, see Fig.~\ref{fig:p2plus_pipline}(middle). However, it  ignores the replacement in the unconditional branch. This leads to less accurate editing in some cases, especially when the structural changes before and after editing are relatively large (e.g., “...tent...”$\rightarrow$“...tiger...” ).  To address this problem, we need to reduce the dependence of the structure on the source prompt and provide more freedom to generate the structure following the target prompt. Thus, as shown in Fig.~\ref{fig:p2plus_pipline}(bottom) we propose to further perform the self-attention map replacement in  the unconditional branch based on P2P (an approach we call \textit{P2Plus}), as well as in the conditional branch like P2P.   This technique provides
more accurate editing capabilities.  Like P2P, we introduce   a timestep parameter $\tau_u$ that determines until which step the injection is applied. Fig.~\ref{fig:uncondselfattn} shows the results with different  $\tau_u$  values.

\section{Experimental setup} 
\subsection{Training details and datasets}
We use the pretrained Stable Diffusion model. The configuration of the mapping network $M_t$ is provided in Tab.~\ref{tab:network}. We set $T=30$. $\tau_v$ is a timestep parameter that determines which timestep is used by the output of the mapping network in the StyleDiffusion editing phase. Similarly, we can set the timestep $\tau_u$ (as in the conditional branch in P2P) to control the number of diffusion steps in which the injection of the unconditional branch is applied.
We use Adam~\cite{kingma2014adam} with a batch size of 1 and a learning rate of 0.0001. The exponential decay rates are $( \beta_{1},\beta_{2})$ $= (0, 0.999)$. We randomly initialize the weights of the mapping network following a Gaussian distribution centered at 0 with 0.01 standard deviation. We use one Quadro RTX 3090 GPUs (24 GB VRAM) to conduct all our experiments.  We randomly collect a real image dataset of 100 images (with resolution $512 \times 512$) and caption pairs from Unsplash(\url{https://unsplash.com/}) and COCO~\cite{chen2015microsoft}.

\subsection{Evaluation metrics}
\textit{Clipscore}~\cite{hessel2021clipscore} is a metric that evaluates the quality of a pair of a prompt and an edited image. To evaluate the preservation of structure information after editing,  we use Structure Dist~\cite{tumanyan2022splicing} to compute the structural consistency of the edited image.  Furthermore, we aim to modify the selected region, which corresponds to the target prompt, while preserve the non-selected region. Thus, we also need to evaluate change in the non-selected region after editing. To automatically determine the non-selected region of the edited image, we use a binary method to generate the raw mask from the attention map. Then we invert it to get the non-selected region mask. Using the non-selected region mask, we compute the non-selected region LPIPS~\cite{zhang2018unreasonable}  between the real and edited images, which we denote \textit{NS-LPIPS} for non-selected LPIPS. A lower NS-LPIPS score means that the non-selected region is more similar to the input image. We also use both PSNR and SSIM to evaluate image reconstruction.

\subsection{Baselines}
We compare our method against the following baselines.
\emph{Null-text}~\cite{mokady2022null}  inverts real images with corresponding captions into the text embedding of the unconditional part of the classifier-free diffusion model.  \emph{SDEdit}~\cite{meng2021sdedit} introduces a stochastic differential equation to generate realistic images through an iterative denoising process.  \emph{Pix2pix-zero}~\cite{parmar2023zero} edits the real image to find the potential direction from the source to the target words. \textit{DDIM with word swap}~\cite{parmar2023zero} performs DDIM sampling with an edited prompt generated by swapping the source word with the target. We use the official published codes for the baselines in our comparison to StyleDiffusion. 
We also ablate variants of StyleDiffusion.

\begin{table}[t]
    \setlength{\tabcolsep}{1mm}
    \caption{Configuration of mappingnetwork $M_t$. $C_I$, $C_O$ denote numbers of input and output channels.}\label{tab:network}
    \resizebox{\columnwidth}{!}{%
    \centering
    \setlength{\tabcolsep}{1pt}
    \begin{tabular}{|c|c|c|c|c|c|}
    \hline
    Metric   & \makecell{(Input channel,\\Output channel)}	&\makecell{(Kernal size,\\Stride)}& \makecell{Input\\dimension}&  \makecell{Output\\dimension} \cr\cline{1-5}
      Conv0  & (197, 77)&(1, 1) &(197, 768)&(77, 768)\cr\cline{1-5}
      Conv1  & (77, 77)&(1, 1)&(77, 768)&	(77, 768)\cr\cline{1-5}
    BatN1  & -&-&(77, 768)&	(77, 768)\cr\cline{1-5}
      LeakyRelu1  & -&-	&(77, 768)&	(77, 768)\cr\cline{1-5}
      Conv2  & (77, 77)&(1, 1)&(77, 768)&(77, 768) \cr\cline{1-5}
    \hline 
    \end{tabular}  
    }
\end{table}

\begin{table}[t]
    \setlength{\tabcolsep}{1mm}
    \caption{Comparison to baselines on three metrics. *DDIM: DDIM inversion with word swap.  Although \textit{DDIM with word swap} achieves the best Clipscore, it not only changes the background, but also modifies the structure of the selected region, as can be seen in Fig.~\ref{fig:cat2dog}(last three rows, fourth column).}\label{tab:scores}
    \resizebox{\columnwidth}{!}{%
    \centering
    \setlength{\tabcolsep}{10pt}
    \begin{tabular}{|c|c|c|c|c|}
    \hline

        Metric   & Structure-dist$\downarrow$	&NS-LPIPS$\downarrow$	&Clipscore$\uparrow$\cr\cline{1-4}

      *DDIM  & 0.092	&0.4131 &\textbf{81.9$\%$ }\cr\cline{1-4}
      SDEdit  & 0.046	&0.2473&78.0$\%$ 	\cr\cline{1-4}
      Null-text  & 0.027	&0.1480	&75.2$\%$ \cr\cline{1-4}
      {MasaCtrl} & 0.039	&0.2259	&80.4$\%$ \cr\cline{1-4}
      Ours  & \textbf{0.026}&\textbf{0.1165}&77.9$\%$ \cr\cline{1-4}
    \hline 
    \end{tabular}  
    }
\end{table}

\begin{table}[t]
    \setlength{\tabcolsep}{1mm}
    \caption{Inference time and PSNR/SSIM. We have better reconstruction quality with a small computational overhead. 
}\label{tab:infer_time}
    \resizebox{\columnwidth}{!}{%
    \centering
    \setlength{\tabcolsep}{10pt}
    \begin{tabular}{|c|c|c|}
    \hline

               & \thead{Inference Time$\downarrow$} & \thead{PSNR$\uparrow$/SSIM$\uparrow$} \cr\cline{1-3}
    Null-text  & \textbf{0.28s}	& 31.314$\slash$0.730 \cr\cline{1-3}
    Ours  & 0.30s & \textbf{31.523$\slash$0.751} \cr\cline{1-3}
    \hline 
    \end{tabular} 
    }
\end{table}

\begin{figure}[t]
    \centering
\includegraphics[width=\columnwidth]{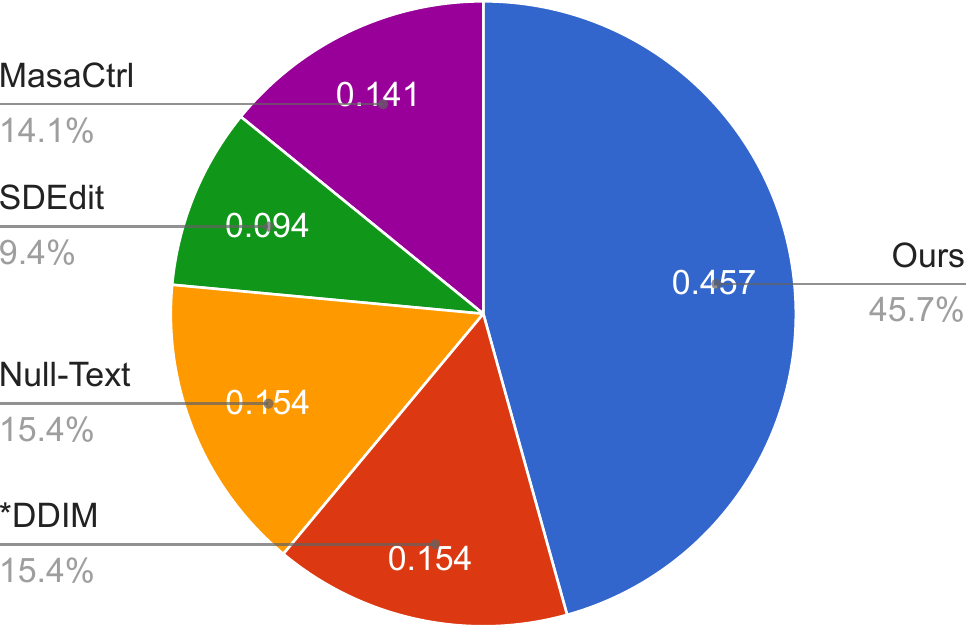}
        \caption{We conduct a forced-choice user study and ask subjects to select the results according to \textit{``Which figure does best preserve the input image structure and matches target prompt style?''}.}
    \label{fig:user_study}
\end{figure}

\section{Experiments}

\subsection{Qualitative and quantitative results}
Fig.~\ref{fig:cat2dog}  presents a comparison between the baselines and our method. \textit{SDEit}~\cite{meng2021sdedit} fails to generate high-quality images, such as dog or cat faces (second column).  
\emph{Pix2pix-zero}~\cite{parmar2023zero} synthesizes better results, but it also modifies the non-selected region, such as removing the plant when translating cat $\rightarrow$ dog.   The official implementation of \emph{pix2pix-zero}~\cite{parmar2023zero} provides the editing directions (e.g, cat $\leftrightarrow$ dog), and we directly use them. Note that \emph{pix2pix-zero}~\cite{parmar2023zero} requires that the editing directions are calculated in advance, while our method does not require this.   Fig.~\ref{fig:cat2dog}(last three rows, fourth column) shows that \textit{DDIM with word swap} largely modifies both the background and the structural information of the foreground. 
MasaCtrl~\cite{cao2023masactrl} is designed for non-rigid editing and tries to maintain content consistency after editing. It often changes the shape of objects when translating from one to another (Fig.~\ref{fig:cat2dog}(fifth column)). For example, when translating dog to cat, although MasaCtrl successfully performs the translation, it changes the shape, such as making the head of the cat larger than in the original image (Fig.~\ref{fig:cat2dog}(second row, the fifth column)).
Our method successfully edits the target-specific object, resulting in a high-quality image, indicating that our proposed method has more accurate editing capabilities.

We evaluate the performance of the proposed method on the collected dataset. Tab.~\ref{tab:scores} reports,  in terms of both Structure distance and NS-LPIPS, that the proposed method achieves the best score, indicates that we have superior capabilities to preserve structural information. In terms of Clipscore, we get a better score than Null-text (i.e., 77.9$\%$ vs 75.2$\%$), and a comparative result with SDEdit. \textit{DDIM with word swap} achieves the best Clipscore. However, \textit{DDIM with word swap} not only changes the background, but also modifies the structure of the selected-regions (see Fig.~\ref{fig:cat2dog}(last three rows, fourth column)). Note that we do not compare to pix2pix-zero~\cite{parmar2023zero} in Fig.~\ref{fig:cat2dog}(last three rows), since it first needs to compute the textual embedding directions with thousands of sentences using GPT-3~\cite{brown2020language}. We also evaluate the reconstruction quality and the inference time for each timestep. As reported in  Tab.~\ref{tab:infer_time}, we achieve the best PSNR/SSIM scores, with an acceptable time overhead.

\begin{figure*}[ht!]
    \centering
\includegraphics[width=\textwidth]{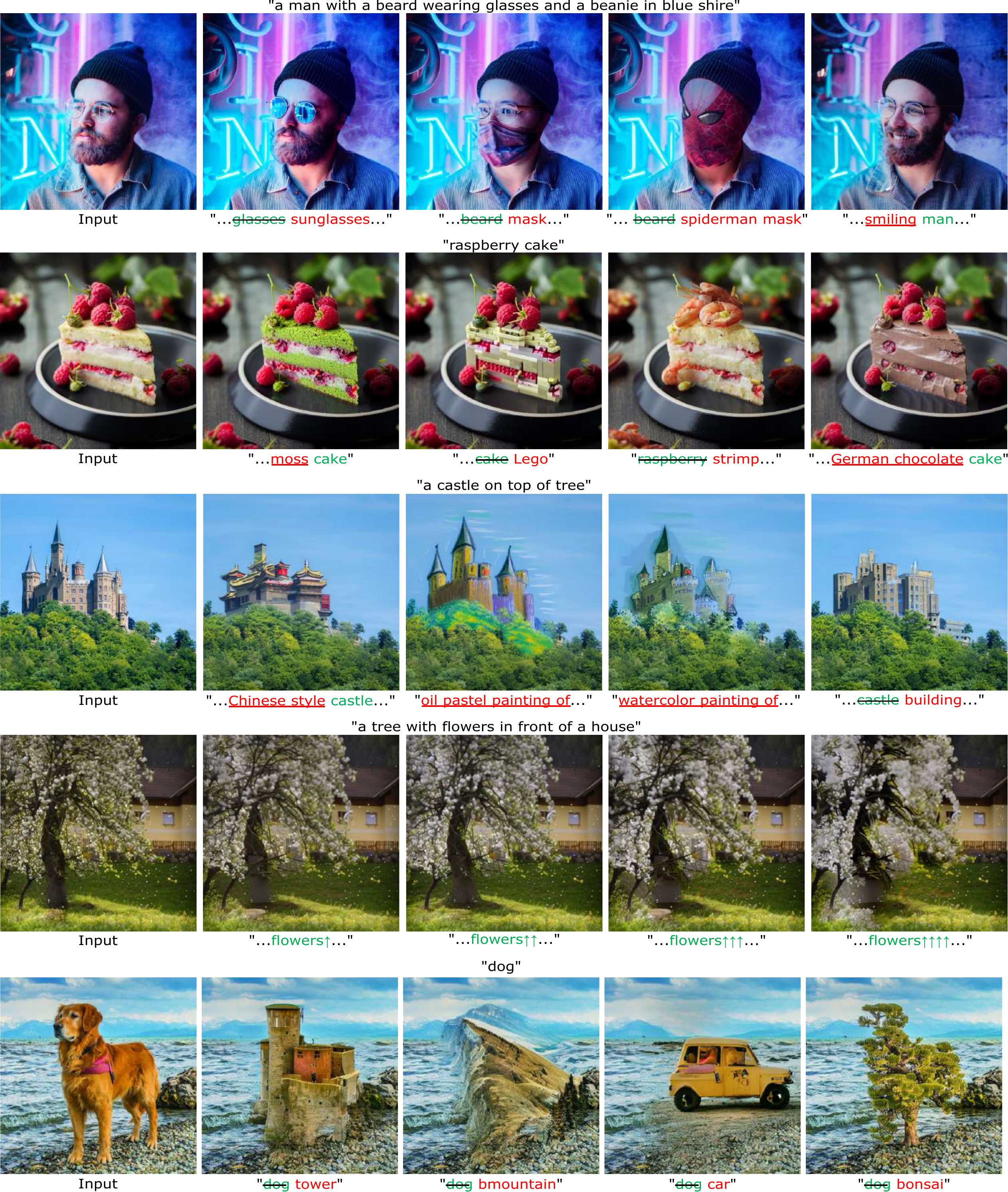} 
        \caption{Examples of StyleDiffusion for editing with 
        attention injection (replacement), refinement (adding a new phrase) or re-weighting.}
    \label{fig:oursresults}
\end{figure*}

\begin{figure*}[t]
    \centering
    \includegraphics[width=\textwidth]{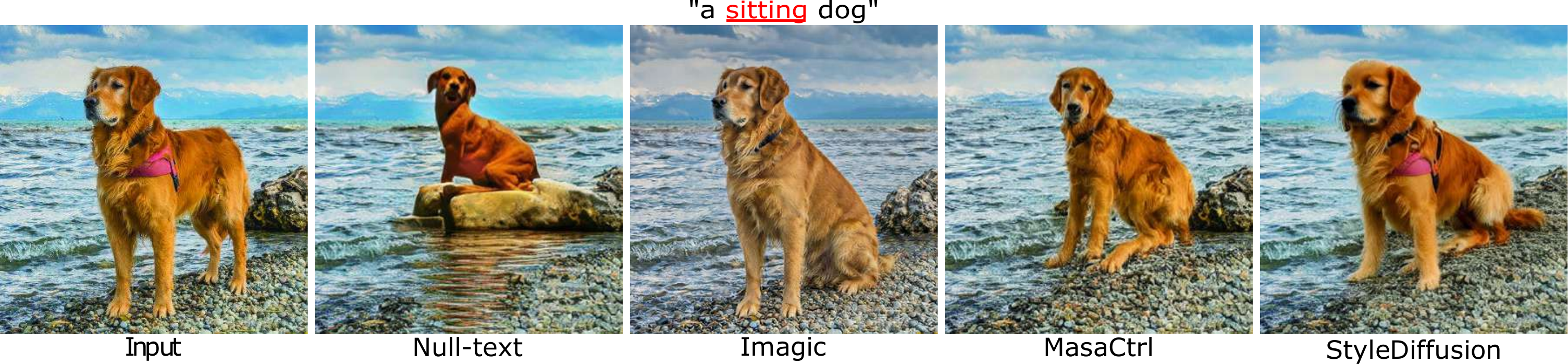}
        \caption{StyleDiffusion can achieve object structural changes within the range of the input image cross-attention map (e.g., “...dog”$\rightarrow$“...sitting dog”).}
    \label{fig:compa3}
\end{figure*}

Furthermore, we conduct a user study, asking subjects to select the results that best match the following statement: \textit{which figure preserves the input image structure and matches the target prompt style} (Fig.~\ref{fig:user_study}). We apply quadruplet comparisons (forced choice) with 54 users (30 quadruplets/user). 
The study participants were volunteers from our college. The questionnaire consisted of 30 questions, each presenting the original image, as well as the results of various baselines and our method. Users were tasked with selecting an image in which the target image is more accurately edited compared to the original image. Each question in the questionnaire presents five options, including baselines (DDIM with word swap, Nul-text, SDEdit, and MasaCtrl) and our method, from which users were instructed to choose one. A total of 54 users participated, resulting in a combined total of 1620 samples (30 questions $\times$ 1 option $\times$ 54 users) with 740 samples (45.68\%) favoring our method. In the results of the user study, the values for DDIM with word swap, NullText, SDEdit, MasaCtrl, and Ours are 15.37\%, 15.43\%, 9.38\%, 14.14\%, and 45.68\%, respectively.

\begin{figure}[t]
    \centering
\includegraphics[width=\columnwidth]{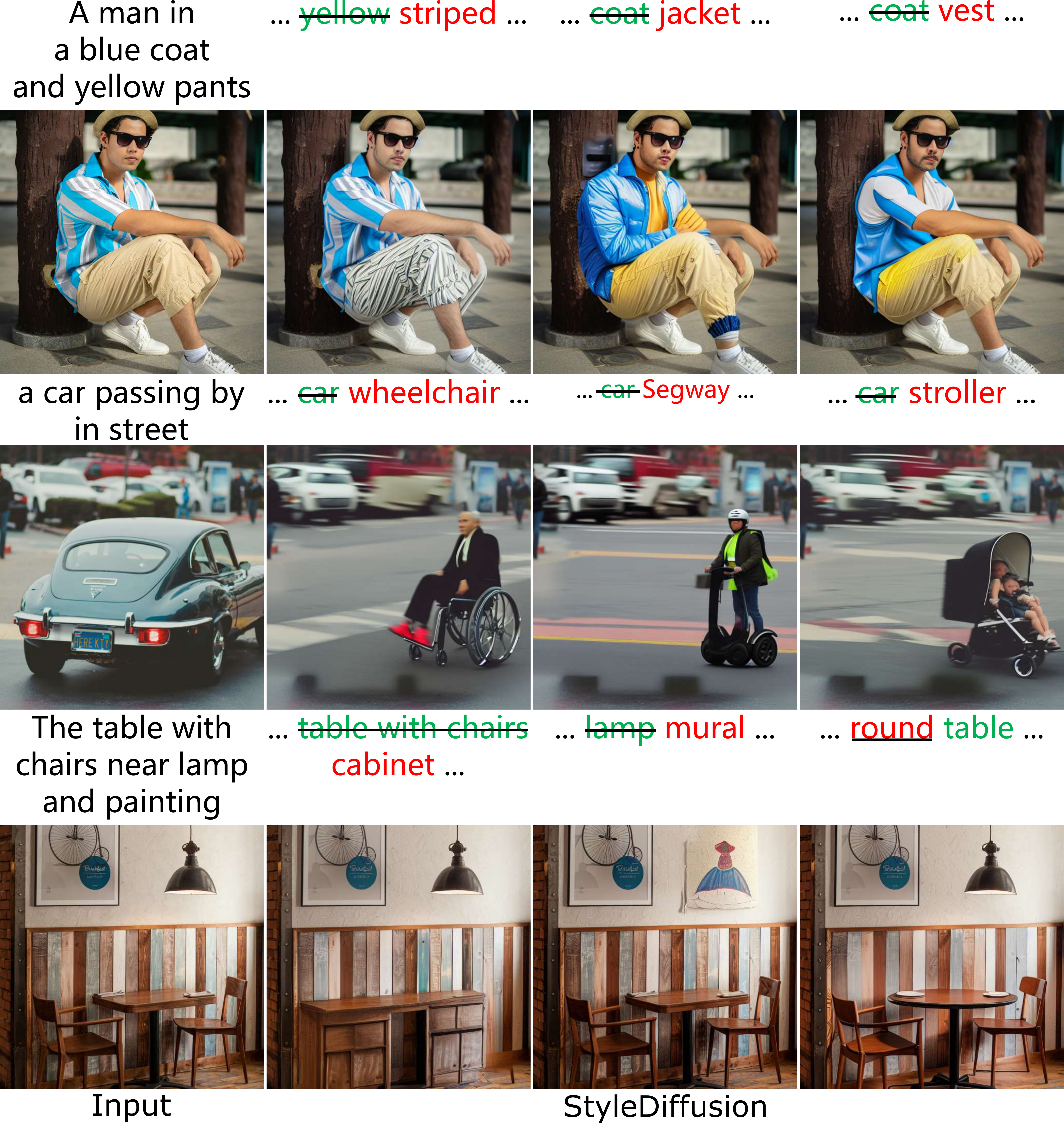}
        \caption{StyleDiffusion can be applied to many real-world applications. For example, in Fashion, rare class insertion for autonomous driving datasets, and interior design.}
    \label{fig:application}
\end{figure}

\begin{figure*}[t]
    \centering
\includegraphics[width=\textwidth]{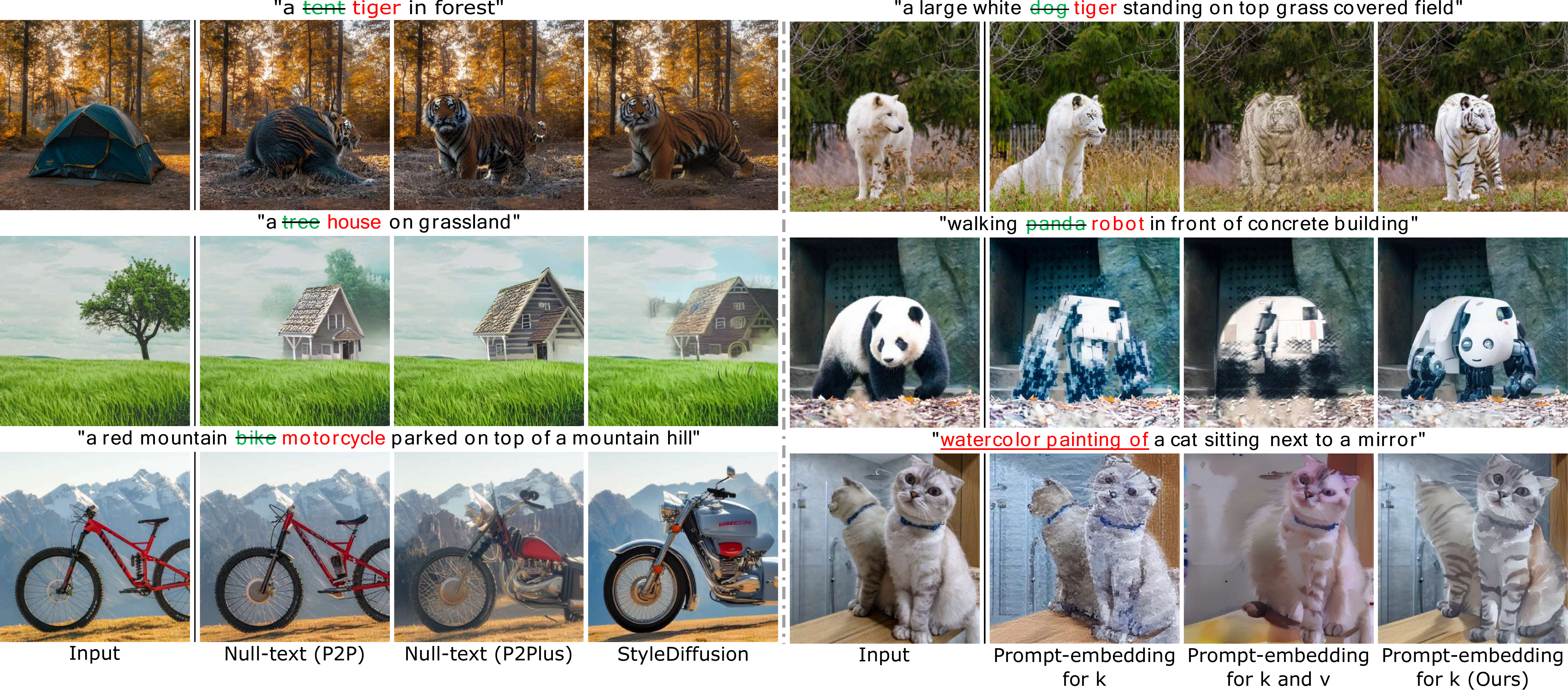}\vspace{-2mm}
        \caption{
        (Left) Additionally using the attention injection in unconditional branch improves the real image editing ability of \emph{Null-text}
        ~(\emph{P2P}). (Right) Comparison of  variants of our method.}
    \label{fig:uncondselfattn_more}
\end{figure*}

\begin{figure*}[t]
    \centering
    \includegraphics[width=\textwidth]{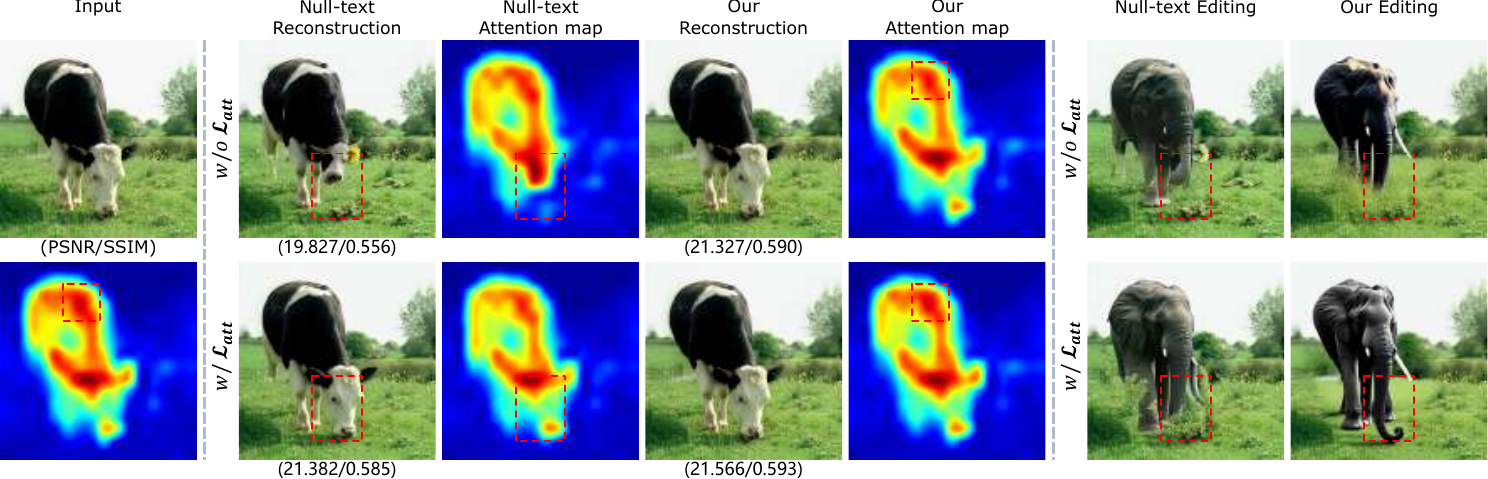}\vspace{-2mm}
        \caption{Reconstruction effect of attention regularization. The cross-attention map of our reconstruction image (second row, fifth column) more closely matches the one of the input image (second row, first column). Meanwhile, $\mathcal{L}_{att}$ can improve the reconstruction quality of Null-text (second row, second column). In the last two columns, the \emph{cow} is replaced by an \emph{elephant} (source prompt: “cow”, target prompt: “elephant”). }
    \label{fig:attnloss}
\end{figure*}

\begin{figure*}[t]
    \centering
    \includegraphics[width=\textwidth]{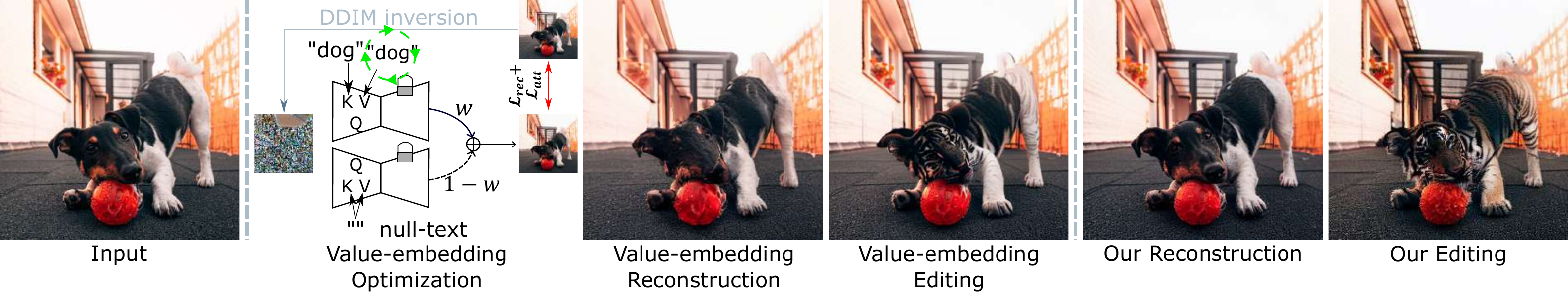}\vspace{-2mm}
        \caption{Value-embedding optimization. We compare our method (right) to the one (middle) in which we directly optimize the input textual embedding for the \textit{value} linear layer while freezing the input of the \textit{key} linear layer. As can be seen (middle), this approach leads to an inaccurate reconstruction, resulting in the dog's face not being completely reconstructed.
        }
    \label{fig:v_embed_opt}
\end{figure*}

Fig.~\ref{fig:oursresults}  shows that we can manipulate the inverted image with 
attention injection (replacement), refinement (adding a new phrase) or re-weighting using P2Plus.
For example, we translate  \textit{glasses}  into \textit{sunglasses} (Fig.~\ref{fig:oursresults}(first row)). We add \textit{Chinese style} (new prompts) to the source prompt (Fig.~\ref{fig:oursresults}(third row)).
We scale the attention map of the ``flowers'' in prompt ``a tree with flowers in front of a house'', resulting in a stronger effect (Fig.~\ref{fig:oursresults}(fourth row)).
These results indicate that our approach manages to invert real images with corresponding captions into the latent space, while maintaining powerful editing capabilities.

We observe that StyleDiffusion (Fig.~\ref{fig:compa3}(last column)) allows for object structure modifications while preserving the identity within the range given by the input image cross-attention map, resembling the capabilities demonstrated by  Imagic~\cite{Kawar2022ImagicTR} 
and MasaCtrl~\cite{cao2023masactrl} (Fig.~\ref{fig:compa3}(third and fourth columns)).
Furthermore, StyleDiffusion can preserve more accurate content, such as the scarf around the dog's neck (Fig.~\ref{fig:compa3}(fifth column)). 
In contrast, Null-text~\cite{mokady2022null} does not possess the capacity to accomplish such changes (Fig.~\ref{fig:compa3}(second column)).

StyleDiffusion can additionally be applied to many real-world applications, including object replacement in images for publicity purposes (as well as personalized advertising: see Fig.~\ref{fig:application}(first row)), dataset enrichment by adding rare objects (e.g., wheelchairs in autonomous driving datasets: Fig.~\ref{fig:application}(second row)), furniture replacement for interior design (Fig.~\ref{fig:application}(third row)), etc.

\subsection{Ablation study.}  
Here, we evaluate the effect of each independent contribution to our method and their combinations.

\subsubsection{Attention injection in the unconditional branch}
Although P2P obtains satisfactory editing results with attention injection in the conditional branch, it ignores attention injection in the unconditional branch (as proposed by our P2Plus in Sec.~\ref{subsec:p2plus}).
We experimentally observe that the self-attention maps in the unconditional branch play an important role in obtaining more accurate editing capabilities, especially when the object structure changes before and after editing of the real image are relatively large, e.g., translating \textit{bike} to \textit{motorcycle} in Fig.~\ref{fig:uncondselfattn_more}(left, third row).  It also shows that the unconditional branch contains much useful texture and structure information, allowing us to reduce the influence of the unwanted structure of the input image.

\subsubsection{Prompt-embedding in cross-attention layers}
We evaluate variants of our method, namely
(i) learning the input prompt-embedding for the \textit{key} linear layer and freezing the input of the \textit{value} linear layer with the one provided by the user, and (ii) learning the prompt-embedding for both \textit{key} and \textit{value} linear layers. As Fig.~\ref{fig:uncondselfattn_more}(right) shows, the two variants fail to edit the image according to the target prompt. Our method successfully modifies the real image with the target prompt, and produces realistic results.

\subsubsection{Attention regularization}
We perform an ablation study of attention regularization. Fig.~\ref{fig:attnloss} shows that the system fails to reconstruct partial object information (e.g., the nose in Fig.~\ref{fig:attnloss}(first row, second column)), and learns a less accurate attention map (e.g., the nose attention map in Fig.~\ref{fig:attnloss}(first row, third column). Our method not only synthesizes high-quality images, but also learns a better attention map even than the one generated by DDIM inversion (Fig.~\ref{fig:attnloss}(second row, first column)).

\begin{figure*}[t]
    \centering
    \includegraphics[width=\linewidth]{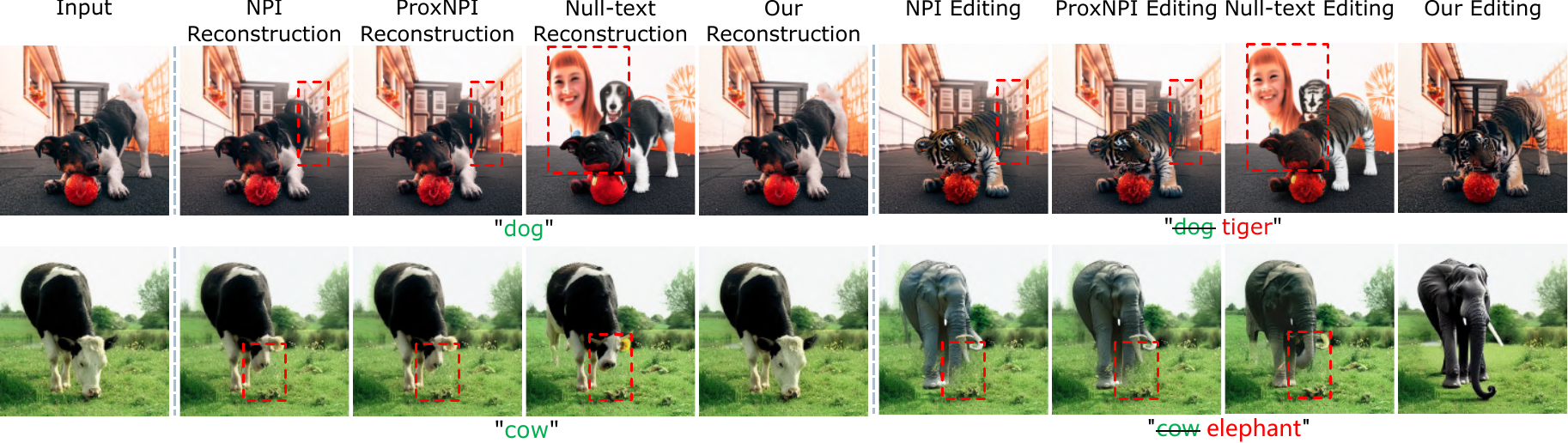}
        \caption{Both the optimization-free methods NPI and ProxNPI (second and third columns, sixth and seventh columns) show limitations in reconstructing and editing real images with complex structures and content.}
    \label{fig:npi}
\end{figure*}

\subsubsection{Value-embedding optimization}
Fig.~\ref{fig:v_embed_opt} illustrates the reconstruction and editing results of value-embedding optimization, that is, similar to our method extracting the prompt-embedding from the input image but directly optimizing the input textual embedding. Value-embedding optimization fails to reconstruct the input image.  Null-text~\cite{mokady2022null} draws a similar conclusion that optimizing both the input textual embedding for the \textit{value} and \textit{key} linear layers results in lower editing accuracy.

\subsubsection{StyleDiffusion with SDEdit}
\begin{figure}[t]
    \centering
    \includegraphics[width=\columnwidth]{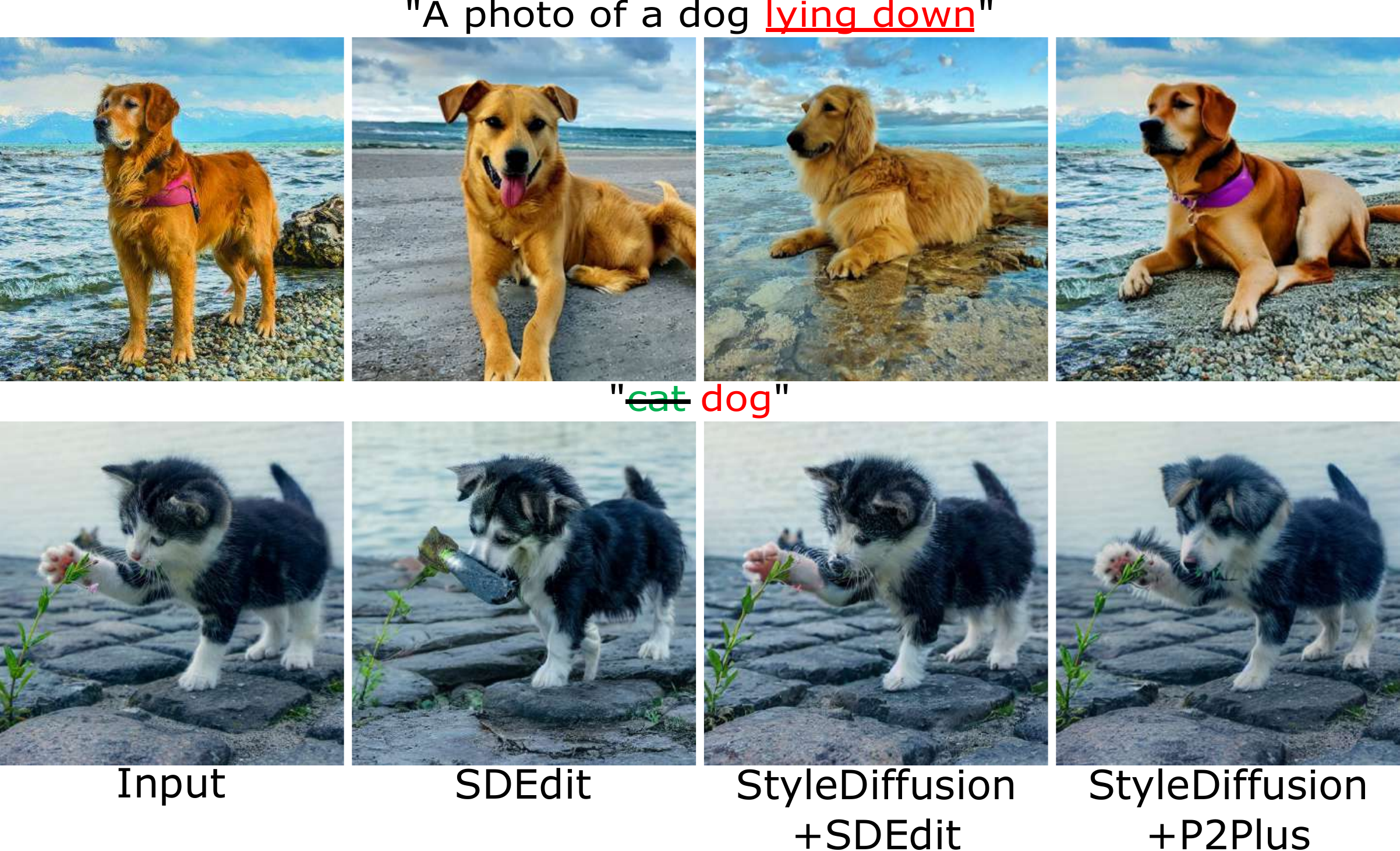}
        \caption{StyleDiffusion with SDEdit. From left to right: input image, SDEdit, applying SDEdit after StyleDiffusion inversion, and applying P2Plus after StyleDiffusion inversion. It is evident that StyleDiffusion+SDEdit significantly improves the fidelity of the input image compared to SDEdit alone.}
    \label{fig:stylediffusion_sdedit}
\end{figure}

\begin{figure}[t]
    \centering
\includegraphics[width=\columnwidth]{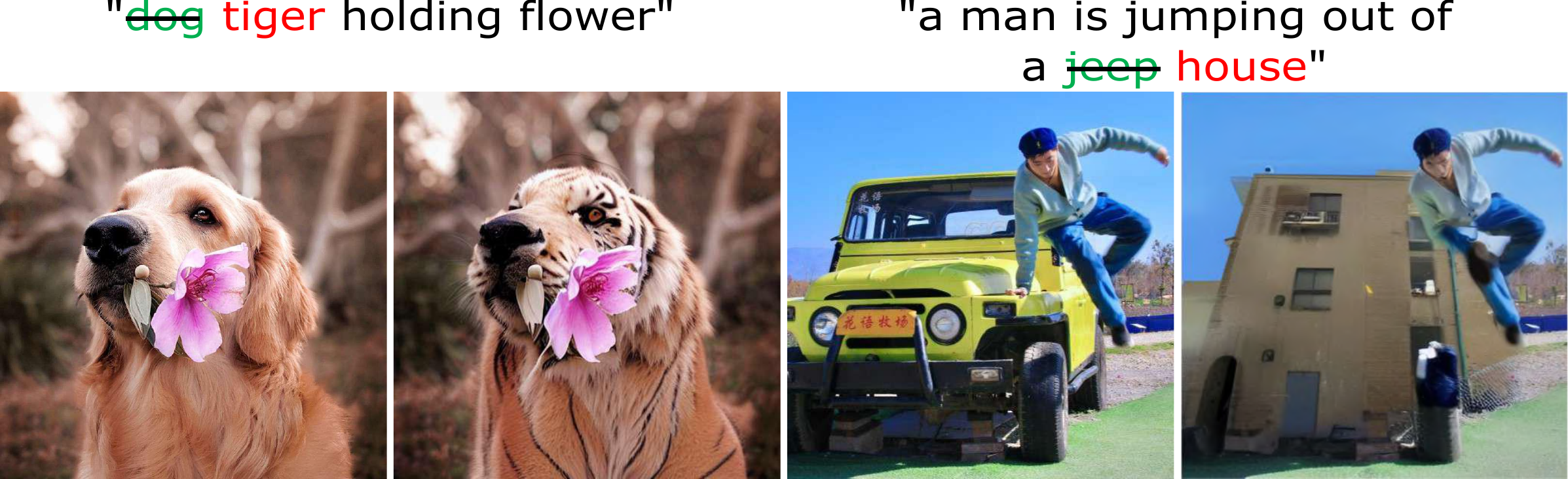}
        \caption{Some examples of failure cases.
        }
    \label{fig:failure_compressed}
\end{figure}

\begin{figure*}[t]
    \centering
\includegraphics[width=\textwidth]{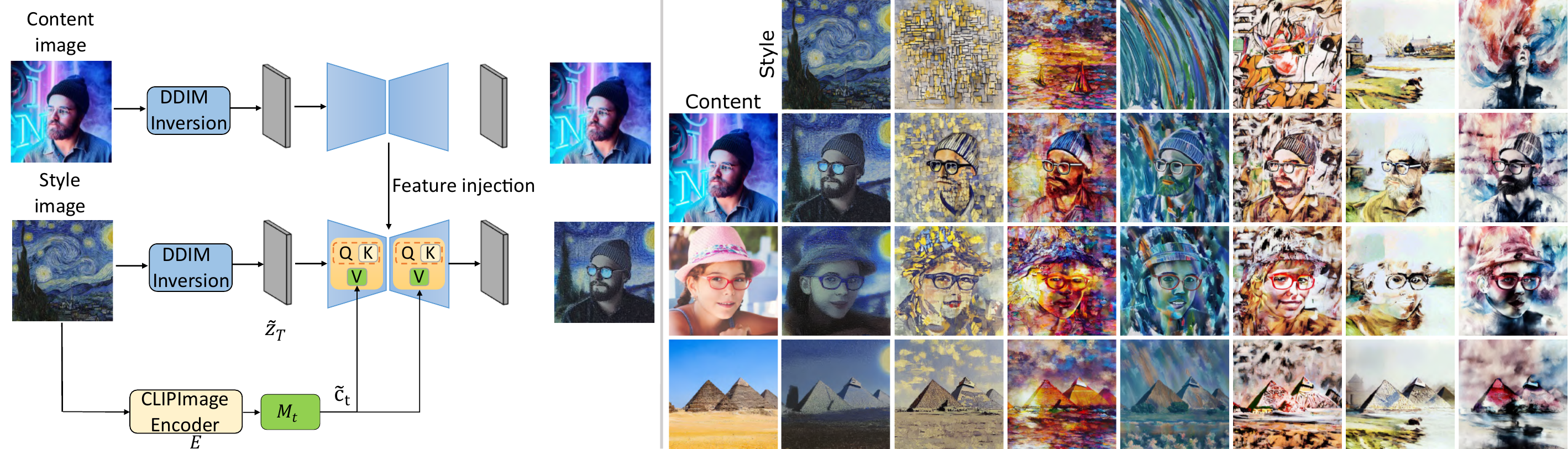}
        \caption{StyleDiffusion for style transfer. (Left) our framework for style transfer. Given a content image, we use DDIM inversion to generate a series of timestep-related latent codes. They are then progressively denoised using DDIM sampling. During this
process, we extract the spatial features from the decoder layers. These  spatial features are injected into the corresponding layers of StyleDiffusion model. Note we first optimize StyleDiffusion to reconstruct the style image, then use both the learned-well $M_t$ and the extracted content feature to perform the style transfer.  
This approach allows us to efficiently transfer the desired artistic style to the content image without the need for additional optimization on the content image. (Right) results of style transfer with StyleDiffusion.
        }
    \label{fig:styletransfer}
\end{figure*}

After inverting a real image with StyleDiffusion, we leverage SDEdit to edit it.  Only using SDEdit, the results suffer from unwanted changes, such as the orientation of the dog (Fig.~\ref{fig:stylediffusion_sdedit}(first row, second column) and the texture detail of the leg of the dog (Fig.~\ref{fig:stylediffusion_sdedit}(second row, second column)).  While combining StyleDiffusion and SDEdit significantly enhances the fidelity to the input image (see Fig.~\ref{fig:stylediffusion_sdedit}(third column)). This indicates our method exhibits robust performance when combining different editing techniques (e.g., SDEdit and P2Plus).

\subsubsection{Comparison with optimization-free  methods}
Recently, some methods have been proposed~\cite{miyake2023negative,han2023improving} that do not use optimization. 
Negative-prompt inversion (NPI) ~\cite{miyake2023negative} replaces the null-text embedding of the unconditional branch with the textural embedding in SD to implement reconstruction and editing.
Proximal Negative-Prompt Inversion (ProxNPI)~\cite{han2023improving} attempts to enhance NPI by introducing regularization terms using proximal function and reconstruction guidance based on the foundation of NPI.
While these methods do not require optimizing parameters to achieve the inversion of real images, like to the method shown in Fig.~\ref{fig:survey}, they suffer from challenges when reconstructing and editing images containing intricate content and structure (see Fig.~\ref{fig:npi}(second and third columns, sixth and seventh columns)).
Due to the absence of an optimization process in these methods, it is not possible to utilize attention loss to refine the attention maps like Null-text+$\mathcal{L}_{att}$ (Fig.~\ref{fig:attnloss}), consequently limiting the potential for enhancing reconstruction and editing quality.

\subsubsection{StyleDiffusion for style transferb}
As a final illustration, we show that StyleDiffusion can be used to perform style transfer. 
Fig.~\ref{fig:styletransfer}(left) shows how,  given a content image, we use DDIM inversion to generate a series of timestep-related latent codes. They are then progressively denoised using DDIM sampling. During this
process, we extract the spatial features from the decoder layers. These spatial features are injected into the corresponding layers of StyleDiffusion model. Note that we first optimize StyleDiffusion to reconstruct the style image, then use both the well-learned $M_t$ and the extracted content feature to perform the style transfer.  Fig.~\ref{fig:styletransfer}(right) shows that we can successfully combine both content and style images, and perform style transfer.

\section{Conclusions and Limitations} 
We propose a new method for real image editing.  We invert the real image into the input of the \textit{value} linear mapping network in the cross-attention layers, and freeze the input of the \textit{key} linear layer with the textual embedding provided by the user. This allows us to learn initial attention maps, and an approximate trajectory to reconstruct the real image.  We introduce a new attention regularization to preserve the attention maps after editing, enabling us to obtain more accurate editing capabilities. In addition, we propose attention injection in the unconditional branch of the classifier-free diffusion model (P2Plus), further improving the editing capabilities, especially when both source and target prompts have a large domain shift. 

While StyleDiffusion successfully modifies the real image, it still suffers from some limitations.  Our method fails to generate satisfying images when the object in the real image has a rare pose (Fig.~\ref{fig:failure_compressed}(left)), or when both the source and the target prompts have a large semantic shift (Fig.~\ref{fig:failure_compressed}(right)).

\subsection*{Declaration of competing interest}

The authors have no competing interests to declare relevant to the
content in this study. \\

\bibliographystyle{CVM}
{\normalsize  \bibliography{CVM2021-VOS}}

\end{document}